\documentclass[journal, ]{IEEEtran}
\pdfoutput=1
\ifCLASSINFOpdf
\else
\fi
\usepackage{graphicx}
\usepackage[table]{xcolor}
\usepackage{bm}
\graphicspath{{Figures/Generated/}}
\usepackage[draft,author={Stefan}]{pdfcomment}
\usepackage{siunitx}
\usepackage{svg}
\usepackage{pdfpages}

\usepackage{pdflscape}
\usepackage[section]{placeins}
\usepackage{longtable}
\usepackage{booktabs}
\usepackage{bm}
\usepackage{enumitem}
\usepackage{subfigure}
\usepackage{multirow}

\usepackage[acronym]{glossaries}
\setacronymstyle{long-short}
\newacronym{ATI}{AT-I}{applications of type I}
\newacronym{ATII}{AT-II}{applications of type II}
\newacronym{BTI}{BT-I}{reactive or reflex-like behavior (low complexity)}
\newacronym{BTII}{BT-II}{cognitive or model-based behavior (high complexity)}
\newacronym{CSE}{C-SE}{capacitive single-ended mode}
\newacronym{CM}{C-M}{mutual-capacitance mode}
\newacronym{ECT}{ECT}{electrical capacitance tomography}
\newacronym{ORLI}{O-RLI}{reflected light intensity}
\newacronym{OTOF}{O-ToF}{time-of-flight}
\newacronym{OTRI}{O-Tri}{triangulation}
\newacronym{OBB}{O-BB}{break-beam}
\newacronym{DOF}{DoFs}{degrees of freedom}

\newacronym{FMCW}{FMCW}{frequency modulated continuous wave}
\newacronym{SAR}{SAR}{synthetic aperture radar}
\newacronym{AUS}{A-US}{ultrasound}
\newacronym{AS}{A-S}{seashell effect}

\newacronym{IR}{IR}{infrared}
\newacronym{HRI}{HRI}{human-robot interaction}
\newacronym{HRC}{HRC}{human-robot collaboration} 
\newacronym{HCI}{HCI}{human-computer interaction}
\newacronym{HMI}{HMI}{human machine interface}
\newacronym{SNN}{SNN}{smart sensor network}
\newacronym{PIPR}{PIPR}{payload inspection and processing robot}
\newacronym{WHAP}{WHAP}{whole-arm proximity}
\newacronym{TTC}{TTC}{time-to-contact}
\newacronym{UAVs}{UAVs}{unmanned aerial vehicles}
\newacronym{GPR}{GPR}{ground penetration radar}
\newacronym{CMC}{CMC}{carbon micro coils}
 
\definecolor{AE}{rgb}{0,0,0}
\definecolor{Reviewer5}{rgb}{0,0,0}
\definecolor{Reviewer2}{rgb}{0,0,0}
\definecolor{Reviewer4}{rgb}{0,0,0}

\usepackage{fancyhdr}

\let\oldlongtable\longtable
\let\endoldlongtable\endlongtable
\renewenvironment{longtable}{\fontfamily{phv}{}\selectfont\oldlongtable} {
\endoldlongtable \fontfamily{\familydefault}\selectfont}

\newcounter{row}
\newenvironment{doclongtable}{
\setcounter{row}{0}\rowcolors{3}{white}{lightgray!30}\fontfamily{phv}{}\selectfont\longtable} {
\endlongtable \fontfamily{\familydefault}\selectfont}

\usepackage{amsmath}
\usepackage{bm}

\makeglossaries

\begin{document}
%
\title{Proximity Perception in Human-Centered Robotics: A Survey on Sensing Systems and Applications }

%
%
%

\author{Stefan Escaida Navarro$^{*}$, 
        Stephan M\"{u}hlbacher-Karrer$^{*}$, 
        Hosam Alagi$^{*}$, 
        Hubert Zangl, 
        Keisuke Koyama, 
        Bj\"{o}rn~Hein, 
        Christian Duriez,
        and~Joshua R. Smith
\thanks{* These authors contributed equally.}
\thanks{Stefan Escaida Navarro and Christian Duriez are with Inria Lille-Nord Europe, France, e-mail: stefan.escaida-navarro@inria.fr}
\thanks{Stephan M\"{u}hlbacher-Karrer is with JOANNEUM RESEARCH ROBOTICS, Institute for Robotics and Mechatronics, Klagenfurt, Austria, e-mail: stephan.muehlbacher-karrer@joanneum.at}
\thanks{Hosam Alagi is with Karlsruhe Institute of Technology, Institute for Anthropomatics and Robotics - Intelligent Process Automation and Robotics Lab (IAR - IPR), Karlsruhe, Germany, e-mail: hosam.alagi@kit.edu}
\thanks{Bj\"{o}rn Hein is with Karlsruhe Institute of Technology, IAR - IPR, Karlsruhe, Germany, e-mail: bjoern.hein@kit.edu and with University of Applied Sciences Karlsruhe, Germany, e-mail: bjoern.hein@hs-karlsruhe.de}
\thanks{Hubert Zangl is with Alpen-Adria-Universit\"{a}t Klagenfurt, Institute of Smart System Technologies, Sensors and Actuators, Klagenfurt, Austria, e-mail: hubert.zangl@aau.at}
\thanks{Keisuke Koyama is with Osaka University, Department of Systems Innovation, Graduate School of Engineering Science, Osaka, Japan, e-mail: koyama@sys.es.osaka-u.ac.jp}
\thanks{Josuha R. Smith is with University of Washington, Allen School of Computer Science and Engineering, Department of Electrical Engineering, Seattle, United States of America, e-mail: jrs@cs.washington.edu}}

\maketitle

\begin{abstract}

Proximity perception is a technology that has the potential to play an essential role in the future of robotics. It can fulfill the promise of safe, robust, and autonomous systems in industry and everyday life, alongside humans, as well as in remote locations in space and underwater. In this survey paper, we cover the developments of this field from the early days up to the present, with a focus on human-centered robotics. Here, proximity sensors are typically deployed in two scenarios: first, on the exterior of manipulator arms to support safety and interaction functionality, and second, on the inside of grippers or hands to support grasping and exploration. Starting from this observation, we propose a categorization for the approaches found in the literature. To provide a basis for understanding these approaches, we devote effort to present the technologies and different measuring principles that were developed over the years, also providing a summary in form of a table. Then, we show the diversity of applications that have been presented in the literature. Finally, we give an overview of the most important trends that will shape the future of this domain.

\end{abstract}

\begin{IEEEkeywords}
Perception for Grasping and Manipulation; Collision Avoidance; Reactive and Sensor-Based Planning; Object Detection, Segmentation and Categorization
\end{IEEEkeywords}

%
\IEEEpeerreviewmaketitle

%
%
%
%

\section{Introduction}
\label{sec:Intro}
\IEEEPARstart{I}{n} the current robotics research landscape, a lot of effort is still dedicated to overcoming the challenges posed by unstructured environments. Areas, such as medicine, health care, agriculture, Industry\ 4.0, and exploration endeavors in space as well as underwater are awaiting to profit from the robotics technologies currently in development. Furthermore, in terms of unstructured environments or situations, one of the main challenges is to develop robotics technologies that enable a safe and reliable interaction with the human. At the same time, the functionality and intelligence provided by the robot system must justify the investment, meaning its autonomous behavior, oftentimes in environments made for humans, must contribute real value. A hallmark of robust and efficient robot behavior is that task execution does not need to be interrupted in the presence of emergent events. A technology that is capable of addressing these challenges is proximity perception. It has been developed over the years, with first, impactful applications being shown in the late 1980s and early 1990s, which were sparked by seminal developments in robot control. Proximity perception is complementary to the main robotics perceptive modalities of vision and touch. Its use is often motivated by closing the perception gap left by occlusions and blind spots as well as by dealing with pose uncertainty of the robot with respect to objects and its environment. Therefore, one of the big challenges in this domain is to find sensor designs that can coexist with the main existing modalities of vision and touch. 

\begin{figure}[t]
    \centering
    \includegraphics[width=0.45\textwidth]{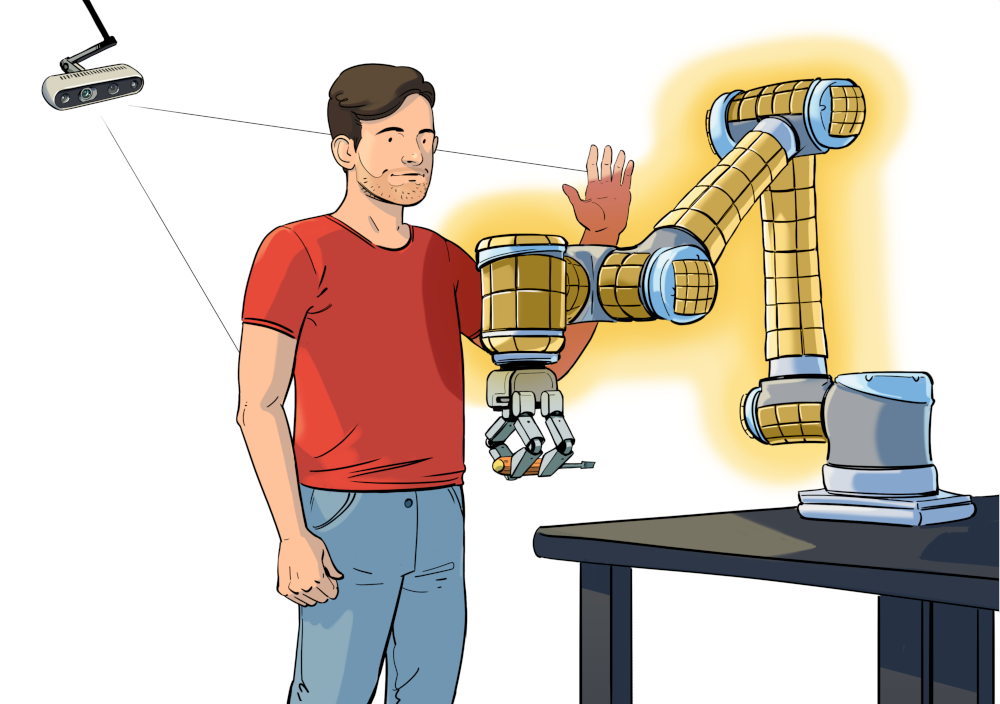}
    \caption{A robot skin with proximity sensing capability can be deployed in scenarios with intense human-robot interaction and collaboration. The skin helps closing the gap between vision-based perception and tactile/force perception.}
\label{fig:OcclusionHRI}
\end{figure}

\begin{figure}
    \centering
    \includegraphics[width=0.45\textwidth]{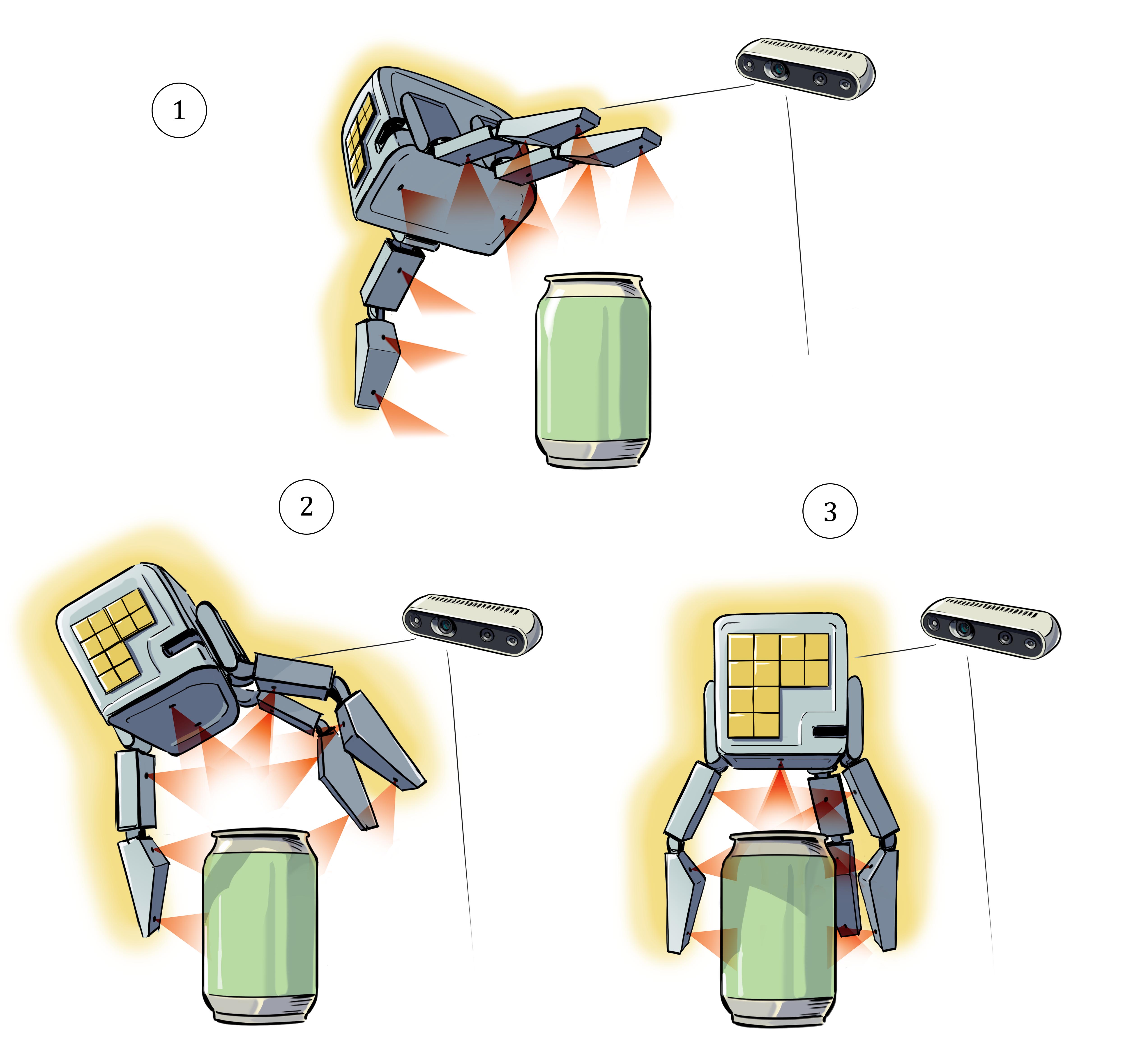}
    \caption{Reactive preshaping to an object based on proximity perception has three characteristic phases: (1) object detection by vision and approaching of the hand to the object, (2) detection of the object by the hand's proximity sensors, start of closed-loop control and occlusion of the object in the camera view, and (3) finalized preshaping control, where the fingers and palm of the hand are aligned with the object.}
\label{fig:Intro_Preshaping}
\end{figure}

In human-centered robotics, the typical applications of proximity perception can be broadly divided into two categories: the first one is pertaining a sensitive skin covering the links of a robot manipulator for safety and interaction functionality, which we call \emph{\gls{ATI}}. The second one is where a robot gripper or hand is equipped with sensors to support grasping and exploration tasks, which we call \emph{\gls{ATII}}. In Fig.~\ref{fig:OcclusionHRI}, a typical scenario of human-robot interaction and collaboration is illustrated (\gls{ATI}). As the human approaches the robot, the view of the camera monitoring the robot and its workspace will become increasingly occluded. A tactile skin covering the robot is not adequate to handle this perception gap in general. This is because detecting the human or the environment only when contact is established, implies operating the robot at very low velocities, thus undermining the purpose of installing such a system in the first place. To address scenarios like these, a sensitive skin with proximity perception capabilities has been proposed by several authors. In Fig.~\ref{fig:Intro_Preshaping}, a typical scenario for grasping supported by proximity perception shown (\gls{ATII}). Since the robotic hand can detect an object's surface before touch is established, a \emph{pre-touch} closed-loop control can be implemented to adjust the hand posture during this phase. This is called \emph{reactive preshaping} and can also have a diversity of use-cases. In Fig.~\ref{fig:Intro_Preshaping}, the three typical phases of this procedure are shown. Proximity perception also has the potential to play an important role in robotic solutions that are compliant with norms and standards, such as ISO/TS 15066 for the operation of collaborative robots.

In this paper, we want to provide an up-to-date perspective on the field of proximity perception in human-centered robotics as well as an introduction to the principles and technologies developed. Proximity perception in areas such as autonomous vehicles or \gls{UAVs} usually aim at autonomous driving or flying and thus address larger distances and speeds and avoidance of contact and interaction with objects and humans. Recent surveys that include discussions on proximity perception in these domains are~\cite{zhou2020mmw} (millimeter wave Radar), \cite{wang2019multi} (sensor fusion), both for autonomous driving, and~\cite{perez2019review} for indoor localization of \gls{UAVs}. In human-centered robotics, proximity perception is related to short distances between humans and robots and aiming for improving human-robot interaction as well as safety. Nonetheless, many ideas presented in this paper can be valid in the automotive domain and for \gls{UAVs} as well, especially those about the sensing principles and their applicability. We hope to give the readers a starting point to understand the principles and applications of proximity perception that have been developed in human-centered robotics. Furthermore, we want to provide a perspective on what, in our opinion, are the important trends that will shape the developments within the next years.

The main contributions of this paper are as follows:
\begin{itemize}
    \item We provide an introduction to the concept of proximity perception and an overview of the possible use-cases. We provide a categorization according to the application types and the complexity of the implemented behavior (Sec.~\ref{sec:ProximitySensor}). We use this categorization throughout the paper to organize the different works found in the literature.
    \item We give an introduction to the working principles of the most important proximity sensor designs (capacitive, optical, radar, etc.) and give a review of the related work in the context of robotics (Sec.~\ref{sec:PhysicalWorkingPrinciples}).
    \item We cover what use-cases for proximity perception have been studied in robotics research and industrial robotics. Here, we give a detailed account of the two most important basic applications, \gls{ATI} and \gls{ATII} (Sec.~\ref{sec:Applications}, see also Figs.~\ref{fig:OcclusionHRI},~\ref{fig:Intro_Preshaping}, and~\ref{fig:ApplicationsAndBehaviors}). We start by giving a historical account and cover the basic forms of behavior possible to the more advanced, cognitive approaches.
    \item We provide a systematic comparison of the technologies from the field summarized in Table~\ref{tab:ComparisonYear}
    \item Finally, we project the current developments into the future and finish the paper with concluding remarks (Sec.~\ref{sec:FuturePerspectives} and \ref{sec:SummaryAndConclusions}).
\end{itemize}

\section{Proximity Sensor: Characterization, Applications and Safety Considerations}
\label{sec:ProximitySensor}

\subsection{Characterization}

\begin{figure}
    \centering
    \footnotesize
    \fontsize{8}{10}\selectfont
    \def\svgwidth{.4\textwidth}
    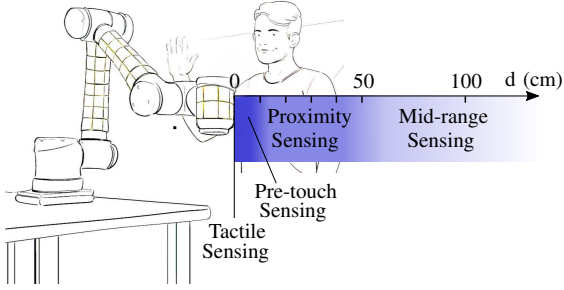
    \caption{Definition of the proximity sensing range.}
    \label{fig:SensingRange}
\end{figure}

Providing a concise characterization of proximity sensors is challenging. One thing common to all proximity sensors is that they detect objects without physical contact. However, this alone does not distinguish them from cameras, which is problematic, as both modalities are considered to be complementary. To address this, we propose a series of attributes that generally characterize proximity sensor designs. At the same time, not all of the attributes need to be present at once in a particular case. Thus, proximity sensors provide non-contact detection of objects \emph{and} more often than not
\begin{itemize}
    \item use active measurement principles, i.\,e. they probe the nearby environment to detect an object's presence, 
    item provide limited sensing range and even small detection ranges can be considered to be useful, 
    \item are \emph{skinlike}, i.\,e. they can be deployed on surfaces such as robot arm segments as well as fingers where they can form a network of sensing elements,
    \item are suitable for being highly integrated into the sensory-motor functionality of the robot, enabling reflex-like behaviors due to low latency measurements,
    \item are used to handle occlusions in vision systems, i.\,e.\ are complementary to vision,
    \item are used to supervise approaching objects which are bound to enter in contact with the robot,  i.\,e.\ are complementary to tactile sensing.
\end{itemize}
In Fig.~\ref{fig:SensingRange}, we propose a definition for the sensing range of a proximity sensor. Any detection distance below $\SI{50}{\centi\meter}$ can be considered to be within the proximity range. This limit is not strict, but in \gls{HRI} and \gls{HRC}, this is an approximate distance at which visual occlusions begin to become problematic. As discussed later in Sec.~\ref{subsec:SafetyConsiderations}, this is a similar range in which monitoring of separation distance is relevant for compliance with safety standards. At larger distances, i.\,e.\ mid-range and long-range perception, other technologies (LIDAR, long-range stereo vision, etc.) can provide better performance in workspace surveillance or for providing \gls{HRI} functionality. This is especially true here because the requirements on reactivity can be relaxed at larger distances. Furthermore, it is interesting to consider the contributions by anthropologist Edward T.\ Hall, who describes the \emph{intimate space} of humans as part of his studies on \emph{proxemics}~\cite{hall1992hidden}. The intimate space starts at a distance of typically $\SI{45}{\centi\meter}$, which is also close to the range proposed above. Therefore, proximity sensing is easy to understand from the perspective of humans, as they can intuitively relate this perception to the ``intimate space" of the robot by analogy. 

\begin{figure}
    \centering
    \fontsize{8}{10}\selectfont
    \def\svgwidth{0.48\textwidth}
    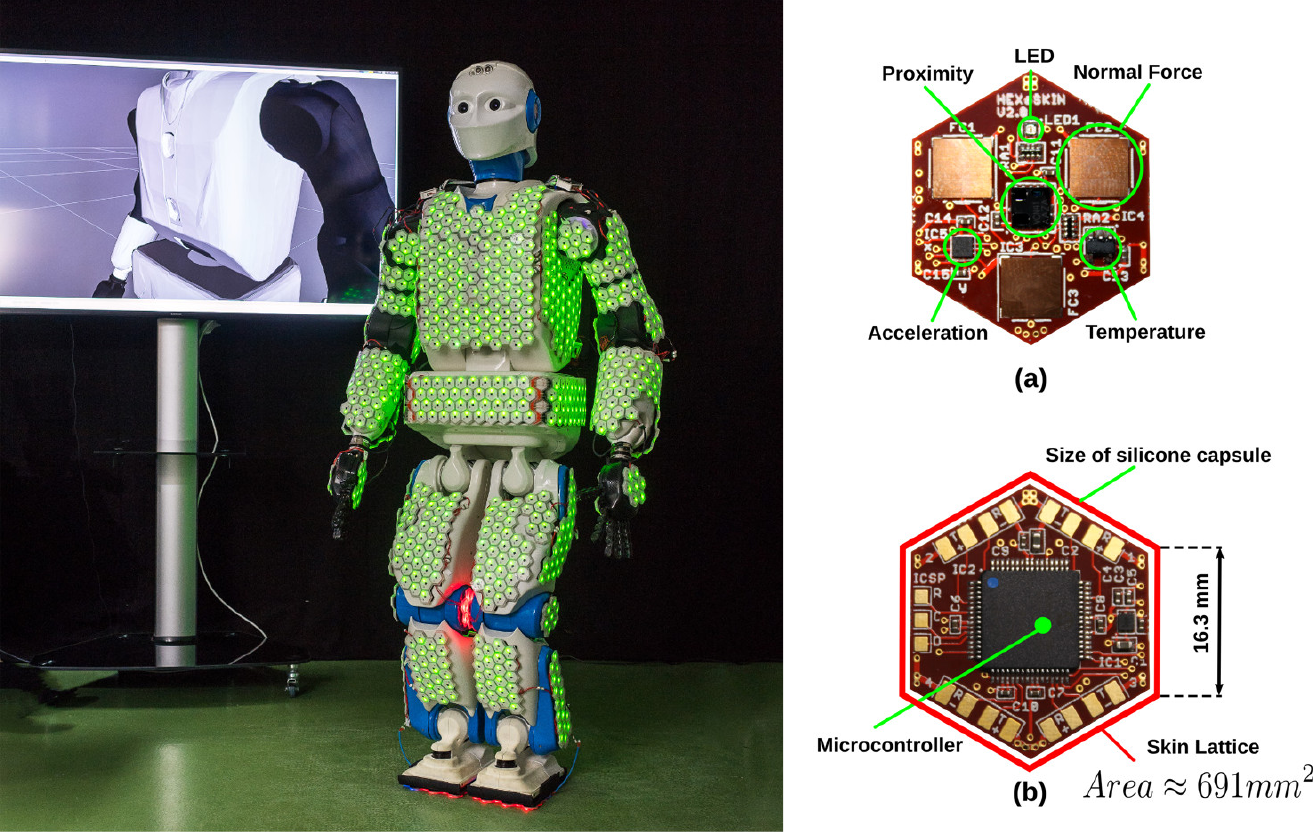
    \caption{Left: Researchers at the Technical University of Munich, Chair for Cognitive Systems, have developed a multi-modal, modular skin, which they have deployed on their robot H-1. Right: A single cell of the multi-modal sensor. (Copyright left picture. A. Eckert / TU M\"unchen. Right picture \copyright 2019 IEEE) \cite{cheng2019comprehensive}. Both reprinted with kind permission of the authors.)}
    \label{fig:LargeScaleModularSkin}
\end{figure}

Finally, a distinction can also be made for a range below $\SI{10}{\centi\meter}$ that we call \emph{pre-touch}-range. This is the type of sensing that precedes contact interactions, for instance during grasping. Here it is especially important to have uninterrupted sensing until contact. Some sensor designs might not feature a long detection range, but the sensing capabilities provided are still useful for closed-loop control of finger and hand posture, which is executed until touch is established. A more in-depth discussion of the available proximity sensing technologies is provided in Sec.~\ref{sec:PhysicalWorkingPrinciples}. In Fig.~\ref{fig:LargeScaleModularSkin}, an example of a modern humanoid robot covered in a multi-modal skin is shown, displaying many of the characteristics discussed in this section.

\subsection{Application and Behavior Types}
\label{subsec:ApplicationsAndBehaviorTypes}
\begin{figure}
    \centering
    \footnotesize
    \includegraphics[width=0.97\linewidth]{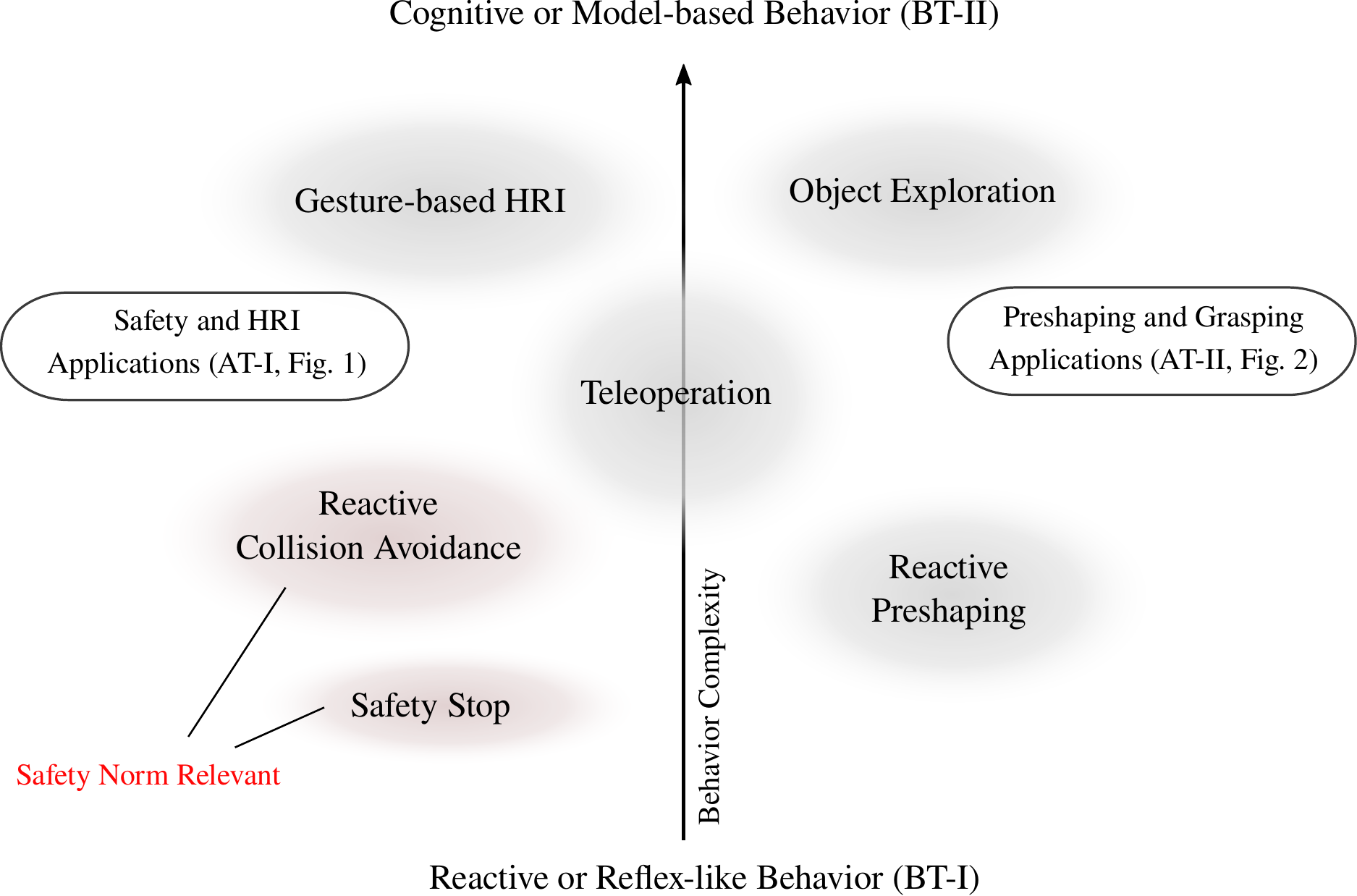}
    \caption{Categorization of some use-cases for proximity sensors in robotics according to the possible application types (\gls{ATI} and \gls{ATII}) and behavior types (\gls{BTI} and \gls{BTII})}.
    \label{fig:ApplicationsAndBehaviors}
\end{figure}

To talk about proximity sensors in human-centered robotics as a whole, it is useful to consider first a categorization of the possible applications and desired behaviors. In the introduction, we already mentioned that a broad classification of applications into two categories is possible: the ones relating to safety and \gls{HRI}/\gls{HRC} and the ones relating to preshaping and grasping (Figs.~\ref{fig:OcclusionHRI} and~\ref{fig:Intro_Preshaping}). Beyond this, automated behaviors based on proximity sensors can be organized according to their conceptual complexity and how instantaneous their effect is on the movement of the robot. One example for a low-complexity behavior is a \emph{safety stop}, i.\,e.\ enabling the brakes of the robot based on a sensor signal surpassing a threshold value. This behavior is closely tied to the update-rate of the sensors and the low-level robot controller. In that sense, it can be called \emph{reactive} or \emph{reflex-like}. Modern collaborative robots, e.\,g.\ the Franka Emika Panda~\cite{FrankaEmika} or the KUKA LBR iiwa~\cite{KUKAiiwa} have control loop cycles of $t_{cl}=1~ms$. Thus, the closer the response time of the proximity sensor is to  $t_r<t_{cl}$, the better. An example of high-complexity behavior is object exploration. It involves managing an object model as well as a planner to complete this model with purposeful exploration steps, resulting in a robot behavior that is executed in several phases and over a longer time compared to the basic control loop cycle times. This behavior is also characterized by being executed at different layers, reaching, as mentioned, up to the planning and cognitive components in the robot's architecture. Fig.~\ref{fig:ApplicationsAndBehaviors} illustrates the categorizing of applications and behaviors we propose as well as providing some examples (not an exhaustive list). As a result, we have a broad classification of applications into two types, \gls{ATI} (left) and \gls{ATII} (right), and behaviors into two types, \gls{BTI} (bottom) and \gls{BTII} (top). In general, \gls{BTI} will appear as subsystems of \gls{BTII}.

\subsection{Safety Considerations and Norm Compliance}
\label{subsec:SafetyConsiderations}
From the safety perspective, a proximity sensor deployed on a collaborative robot in an industrial environment has to fulfill the requirements of the ISO/TS 15066 complying with the ISO 10218 for robots and robotic devices, as well as the performance level defined in the safety of machinery ISO 13849. Proximity sensing hardware is very well suited to operate a collaborative robot in the \emph{speed and separation monitoring mode} as defined in the ISO/TS 15066 to monitor the protective separation distance $S_p$ between a human and a robot's surface. According to the standard, the following equation has to be fulfilled during the operation mode:

\begin{equation}
  \label{eq:separationDistance}
  \small
  \begin{aligned}
    S_{p}(t=t_0) = v_{h} (T_{r}+T_{s}) + v_{r} T_{r} + S_{s} + C_{i} + Z_{d} +Z_{r},\\
  \end{aligned}
\end{equation}
where $v_h$ is the human speed (if not monitored, $v_h=
\SI{1.6}{\meter\per\second}$)
$T_r$ is the reaction time of the robot, $T_s$ is the robot stopping time, $v_r$ is the robot speed, $S_s$ is the robot stopping distance, $C_i$ is the intrusion distance, $Z_d$ is the position uncertainty of the human and $Z_r$ is the position uncertainty of the robot.

To provide an illustrative example for a state of the art collaborative robot, we look at the UR10e series. It has a control loop cycle of $\SI{500}{\hertz}$ and the safety parameters can be configured $v_r =\SI{5}{\meter\per\second}$ (max end-effector speed)
$T_r = \SI{4}{\milli\second}$ (two control loop cycles), $T_s = \SI{100}{\milli\second} $, $S_s = \SI{50}{\milli\meter}$, $Z_r = \SI{0.05}{\milli\meter}$. This configuration results in separation distance of $S_p=\SI{0.236}{\meter}$ excluding the uncertainty of the position of the human and the intrusion distance, as this depends on the sensor parameters monitoring the area.

\section{Measurement Principles for Proximity Sensing}
\label{sec:PhysicalWorkingPrinciples}

In this section, we will give an introduction to the main physical principles available to implement proximity sensing. The idea is to be able to ease the process of reviewing articles by starting with an explanation of the basics. This will also help us in the systematization in Table~\ref{tab:ComparisonYear}. There, we use the abbreviations introduced in this section. Here, we will already do a review of some representative works in the field focusing on how the proximity sensing technology is implemented. This section is closely linked to Sec.~\ref{sec:Applications}, where the details of the implemented applications are discussed. We try to cross-reference the most relevant relationships. However, favoring readability, cross-referencing is not exhaustive.

\subsection{Capacitive Sensing}
\label{sec:CapacitiveSensing}
In this section, we provide a short introduction to capacitive sensing, its measurement techniques as well as the work done dedicated to capacitive proximity sensing in robotics. The capacitive measurement principle has been widely adopted in various other fields and has well-established applications in research and industry, \cite{baxter_capacitive_1997} gives an overview on basic principles and applications. A recent survey paper reviewing capacitive sensing for \gls{HCI} is due to Grosse-Poppendahl et al.~\cite{grosse2017finding}. In this section, we will concentrate on the technologies related to robotics. 

The capacitive proximity sensing principle uses electrically conductive elements (electrodes) to generate and measure electric fields. Objects interfere with this electric field when they approach the electrodes and the observed changes are utilized to estimate their distance as well as properties of the object, such as its material. Therefore, capacitive sensing is called \emph{electric field sensing} in some literature. Essentially, the capacitance between the sensor and an object depends on the geometry of an object, its relative pose to the electrode(s), its coupling to electrical ground, and its material. The nonlinear relation between the relative pose and the material of the object to the measured signal presents a significant challenge for developing signal processing for and applications based on capacitive sensing. However, its ubiquitous use for \gls{HCI} is explained by the fact that humans can be detected reliably.

Commonly, alternating electric potentials are used to generate the electrical field and displacement currents that are proportional to the capacitances are measured. Another popular approach is measuring the oscillation frequency in an oscillator-circuit based on the capacitance of interest. Typically, the alternating frequency is rather low, i.\,e. not much larger than $\SI{1}{\mega\hertz}$, and thus the corresponding wavelength is long compared to the size of the electrodes such that wave-propagation effects can be neglected and the quasi-static assumption can be used. 

\begin{figure}
    \centering
    \footnotesize
    \def\svgwidth{.48\textwidth}
    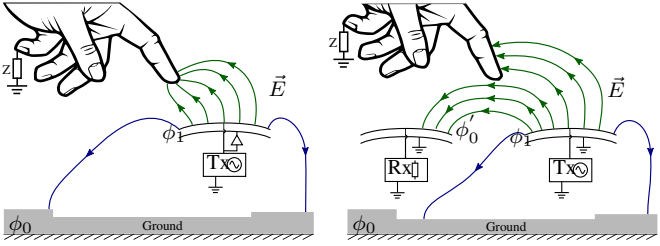
    \caption{\Acrfull{CSE} (left): The electrode is driven with an electrical potential and generating an electrical field between the measurement electrode and the object. \Acrfull{CM} (right): the right electrode transmitter (Tx) is driven, generating an electrical field, which ends at receiver (Rx). A conductive object within the measurement range can block/shield the field lines (green) between the electrodes.}
\label{fig:CapacitiveSensing}
\end{figure}

Mainly two different modes of operations are distinguished for capacitive sensors: The first mode, called \emph{\gls{CSE}}, uses the influence of an object on the capacitance between sensor electrodes and distant ground (see Fig.~\ref{fig:CapacitiveSensing}, left). This mode is also called \emph{self-capacitive mode} or \emph{shunt mode} in literature. The second mode, called \emph{\gls{CM}}, sometimes also \emph{differential mode}, uses the influence of an object on the capacitances between electrodes of the sensor (see Fig.~\ref{fig:CapacitiveSensing}, right).
Both modes are widely used. An advantage of the single-ended mode is a typically higher capacitance and thus a higher signal to noise ratio. The mutual capacitance mode has the advantage of providing more independent measurements, as all the combinations between electrodes can be measured. Therefore, \gls{ECT} (e.\,g.~\cite{MuehlbacherKarrer2016a}) that allows obtaining images of material distributions usually utilize the latter, sometimes in combination with the single-ended mode.   

Fig.~\ref{fig:CapacitiveSensing} (left) shows the self-capacitance mode and illustrates the electrical field between a transmitter electrode and an object. Some electrical field lines end in the ground representing unwanted coupling to the environment, also called parasitic effects. A second layer can be used as an active guard, which has the same electrical potential as the transmitter. This is a popular method to reduce parasitic effects and to actively shape the measurement lobe of the electrode. 
In this mode, approaching objects in general increase the measured capacitance. Conductive objects with strong coupling to ground show a strong sensing effect. Having reasonable conductivity and body parts that offer large areas, humans behave like such objects and are thus well recognized, achieving a high coupling capacitance in the range of $\SI{100}{\pico\farad}$. Small conductive objects and non-conductive objects show lower sensing effects and are more difficult to detect. In case that the sensors are intended to detect humans, the low sensitivity with respect to small objects can be an advantage as this implies that contamination of the electrode surface e.\,g. with water due to condensation or accumulation of dirt will have little impact on the sensor performance.

Fig.~\ref{fig:CapacitiveSensing} (right) shows the mutual capacitance mode and illustrates the electrical field between a transmitter electrode, an object, and a receiver electrode. Conductive objects with high coupling to ground (such as humans) can partially shield the field and thus -- in contrast to the single-ended mode -- reduce the capacitance between the electrodes. If coupling to ground is low, then the opposite effect can occur and the coupling increases due to polarization within the object. This is in particular common for non-conductive objects as well as for higher frequencies of the excitation frequencies. In mutual capacitance mode, shaping of the field can be achieved using passive ground electrodes.

Using the electrostatic representation, the relation between charges $Q$ on the electrodes and the potentials $\Phi$ on the electrodes can be described as
\begin{equation}
\scriptstyle
\begin{pmatrix}
Q_1\\
\vdots  
\\ 
Q_n
\end{pmatrix}
=\begin{pmatrix}
C_{1,1} &\cdots  & -C_{N,1}\\ 
 \vdots&\ddots  & \vdots \\ 
-C_{1,N} & \cdots & C_{N,N}
\end{pmatrix}
\begin{pmatrix}
\Phi_1\\
\vdots  
\\ 
\Phi_n
\end{pmatrix}=\displaystyle{\bm{C}}\begin{pmatrix}
\Phi_1\\
\vdots  
\\ 
\Phi_n
\end{pmatrix}
\end{equation}

where $C_{i,j}$ represents the capacitance between electrode $i$ and electrode $j$ and the diagonal elements represent the capacitances between the electrode $i$, ground (reference potential) and all other electrodes. The self-capacitance mode typically determines the diagonal elements, whereas the mutual capacitance mode determines off-diagonal elements. To perform the inversion and obtain the capacitances, linearly independent excitation patterns are needed to determine the full matrix $\bm{C}$.

\subsubsection{Early Capacitive Technologies in Robotics}

One of the first applications of proximity (and tactile) sensor for robots goes back to 1988 presented by Yamada at el.~\cite{Yamada1988}. Two links of a manipulator are equipped with mutual-capacitance sensors (\gls{CM}) and the capability of detecting conductive and insulating approaching obstacles is demonstrated. It is established that conductive objects are detected more reliably than non-conductive objects. In the 1990s, with the same motivation of avoiding obstacles, other groups worked on capacitive sensing, for instance focusing on the electrode design, like Vranish et al.\ in \cite{vranish_capaciflector_1991}. This technology was evaluated for use in collision avoidance by Wegerif et al.~\cite{wegerif1993whole}, but was dropped in favor of \gls{IR} sensing (see also Secs.~\ref{subsubsec:RLIEarly} and~\ref{subsubsec:EarlyJacobian}). However, authors like Novak and Feddema favored capacitive sensing for these kinds of approaches, developing large sensor arrays to cover greater areas on robot arms \cite{Novak1992} \cite{Feddema1994} (see also Sec.~\ref{subsubsec:EarlyJacobian}). 

\subsubsection{Capacitive Sensing for \gls{ATI}}

Since the first developments, many groups in the robotics community have worked on capacitive based proximity sensing. On the one hand, capacitive sensing has been further investigated to cover robot links (\gls{ATI}, see Sec.~\ref{subsec:CollisionAvoidance}). In \cite{Schlegl2013b} and \cite{MuehlbacherKarrer2016a}, the authors propose a mutual-capacitance sensor (\gls{CM}) having several electrodes for detecting obstacles on robot links. In~\cite{Schlegl2013b}, the ability to detect non-conductive materials due to the mutual-capacitance sensing principle is highlighted. Collision avoidance on a mobile robot based on capacitive proximity sensing for a variety of materials was shown in~\cite{MuehlbacherKarrer2015}. Covering robot links with modular single-ended capacitive proximity (\gls{CSE}) skins for collision avoidance and \gls{HRI} is proposed in~\cite{Stiehl2006,Escaida2016a,ding2019with,tsuji2020proximity}. These works show the potential for these technologies, especially for \gls{HRI}. However, for example in~\cite{Escaida2016a}, it is discussed that \gls{CSE} is not suitable enough to detect insulating materials for the intended application of collision avoidance. As a solution, in~\cite{ding2019with,tsuji2020proximity}, a combination with \gls{OTOF} sensing (see Sec.~\ref{subsubsec:ToF}) is proposed to compensate for the shortcomings of \gls{CSE} sensors. Capacitive proximity sensing has also been adopted by some robotics companies to implement safety-features, especially in \gls{HRI}. Examples are BOSCH APAS \cite{bosch_apas_nodate,Frangen2010}, FOGALE Robotics~\cite{FOGALErobotics,mcolo2019obstacle} as well as MRK-Systeme~\cite{hoffmann2016environment} 
(see Sec.~\ref{subsec:IndustrialTechnologies}).

\subsubsection{Capacitive Sensing in \gls{ATII}}

While the first developments in capacitive proximity sensing in robotics were focused on \gls{ATI}, more recently, \gls{ATII} have started to gain interest. Smith et al.\ showed the use of multi-transmitter and receiver in robotic grippers \cite{smith2007,Wistort2008,Mayton2010}. The placement of transmitter and receiver electrodes allows covering diverse sensing ranges for the different aspects of reactive preshaping, i.\,e.\ short, mid, and long-range sensing for adjusting the poses of the fingers and the palm. Furthermore, they show it is possible to detect the grounding state of an object as it changes due to a human holding on to it. G\"oger et al.\ show the integration of a tactile proximity sensor into fingertips~\cite{Goeger2013}.  In~\cite{Escaida2014b,Escaida2015b}, Escaida~Navarro et al.~show the integration of the sensor presented in~\cite{Goeger2010a} into a two-jaw gripper for reactive preshaping and telemanipulation with force-feedback. In~\cite{Escaida2014b,Escaida2015b}, the limitations of self-capacitance sensing with regards to material properties are on display. This is addressed to some extent in~\cite{Escaida2016b} with mutual-capacitance sensing and flexible spatial resolution. In~\cite{MuehlbacherKarrer2015b}, sensors are integrated into the fingers of a humanoid robot, which help in finding an object's fill state. These works have shown the feasibility of integrating capacitive sensors into the fingertips of robot hands. However, the reduced size of the electrodes remains a challenge. A smaller size is desirable for integration and spatial resolution but is attained at the cost of reduced electrode surface area, which limits the possible sensing range/sensitivity. A further interesting use-case for capacitive sensing is introduced by Erickson et al.\ in~\cite{erickson2018tracking,erickson2019multidimensional}. Using off-the-shelf electronics (MPR121 and the Teensy-board respectively), they implement a capacitive end-effector for the PR2 that is capable of detecting human limbs for dressing and washing tasks in health-care scenarios. The mechanical robustness of capacitive sensors also makes them suitable for harsh industrial environments, like investigated in \cite{Faller2019} for a grasper of an autonomous forestry crane.

\subsubsection{Further Aspects of Capacitive Proximity Sensing Technologies in Robotics}

As the capacitive measurement principle is suitable for implementing both tactile and proximity sensors, there have been efforts to realize both modalities in a single sensor design, e.\,g.\ \cite{Lee2009,Goeger2010a,Goeger2013,han2016highly,Alagi2016a}. This is a special case of multi-modal sensors (see Sec.~\ref{subsec:Multi-ModalSensors}). Other works have made use of the material dependency and investigated material recognition using multi-exciter frequencies \cite{kirchner2008capacitive}, \cite{ding2018capacitive}, and \cite{alagi2018material}. Moreover, tomographic measurements using capacitive sensors were also studied including side effects and material dependencies \cite{MuehlbacherKarrer2015, MuehlbacherKarrer2015f}  and the potential for flexible spatial resolution was explored \cite{Alagi2016a,Escaida2016b,alagi2020}. The possible applications are further discussed in Sec.~\ref{sec:Applications}.

In contrast to optical proximity sensors (see Sec.~\ref{subsec:OpticalSensing}), it is not as common for researchers to use off-the-shelf solutions for implementing capacitive proximity sensors. More often than not, capacitive sensor circuits have been developed by the robotics researchers themselves. Another difference to optical sensing is the attainable sensing rate. The typical rates reported fall in the range of 20-$\SI{125}{\hertz}$ (with some exceptions up to several kilohertz) for capacitive sensing, whereas recent optical sensing approaches report update rates $>\SI{1}{\kilo\hertz}$ (see Sec.~\ref{subsec:OpticalSensing}). The difference can be explained by the fact that the effect an object has on an electric field is often quite weak, leading to low signal to noise ratios and the comparatively low-frequency carrier frequencies for the measurement circuitry. Stronger excitation signals might compensate for the low sensitivity but this is limited due to increasing costs and higher power consumption. Also, often a single sensing front-end is addressing several sensing elements in a time-multiplexed manner, further decreasing the update rate. 

\begin{figure}
    \centering
    \subfigure[\label{fig:ElectricFishAnimal}]{
    \includegraphics[width=0.48\linewidth]{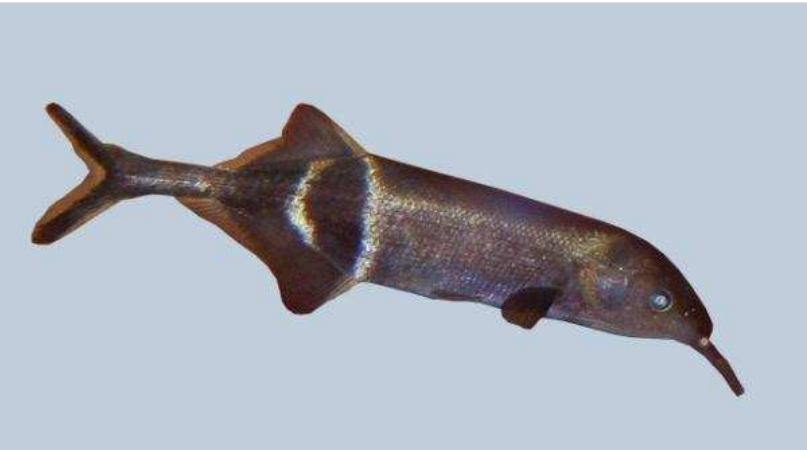}
    }
    \subfigure[\label{fig:ElectricFishField}]{
    \includegraphics[width=0.45\linewidth]{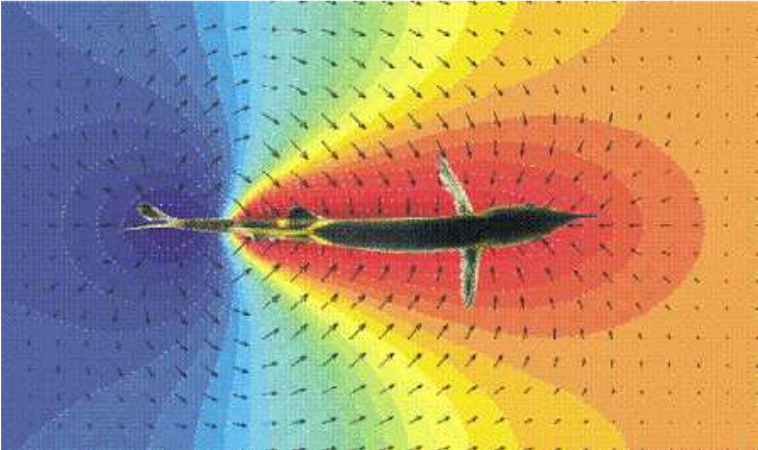}}
    \caption{\subref{fig:ElectricFishAnimal} The \emph{freshwater elephantfish} native to Africa feature an electric organ and electroreceptors, creating a mutual impedance system, for sensing their surroundings, and for communication. This makes them attractive subjects for studying bio-inspired proximity sensing approaches.~\subref{fig:ElectricFishField} An illustration of the electric field generated by the freshwater elephantfish \cite{boyer2018electric}.(Sensing type: similar to \gls{CM}), images courtesy of Fr\'ed\'eric Boyer and Vincent Lebastard \copyright 2013 IEEE)}
\label{fig:ElectricFish}
\end{figure}

A somewhat unique domain in robotics, where capacitive-like sensing plays an important role, is in bio-inspired underwater robots. Here, mimicry of weakly electric fish, that is, fishes that use this sensing modality for navigation, preying, and communication, is studied (see Fig.~\ref{fig:ElectricFish}). The electric fish live in low-visibility and cluttered environments where the \emph{electrosense} becomes a crucial tool. They use an electric organ to generate voltage pulses or oscillations and have voltage receptors on their skin to detect disturbances of the field, i.\,e.\ they implement a mutual impedance system in water similar to a capacitive system (\gls{CM}) in air. Examples are the research by Boyer and Lebastard et al.~\cite{boyer2013underwater} as well as MacIver et al.~\cite{bai2015finding}. Both groups have published an important number of articles on this research topic. 

\subsubsection{Spatial Resolution and Sensing Range for Capacitive Sensors}

Regarding spatial resolution, capacitive sensing is highly adaptable. Reducing the electrode size is not necessarily a problem in terms of fabrication, but sensitivity becomes more challenging as the size decreases. While using cells of size $1\!\times\!\SI{1}{\square\milli\meter}$ for tactile sensing is no problem~\cite{Lee2009}, an area of $\approx 15\!\times\!\SI{25}{\square\milli\meter}$ is needed for detecting conductive objects at a distance of about $\SI{40}{\milli\meter}$ for a sensor mounted on a finger (\gls{ATII}) in \cite{Goeger2013}. However, sensing range is also determined by the distance of the electrodes in mutual-capacitive mode~\cite{Mayton2010} (\gls{ATII}) as well as the circuit design. In~\cite{mcolo2019obstacle}, a detection distance of about $\SI{30}{\centi\meter}$ for electrodes of size between $50\!\times\!\SI{50}{\square\milli\meter}$ and $100\!\times\!\SI{100}{\square\milli\meter}$ is reported for \gls{HRI} (\gls{ATI}). Finally, as the sensing range increases, self-influence becomes an issue that needs to be addressed~\cite{Yamada1988,poeppel2020robust}.
\subsection{Optical Sensing}
\label{subsec:OpticalSensing}

\begin{figure*}
    \centering
    \fontsize{8}{10}\selectfont
    \def\svgwidth{0.75\textwidth}
    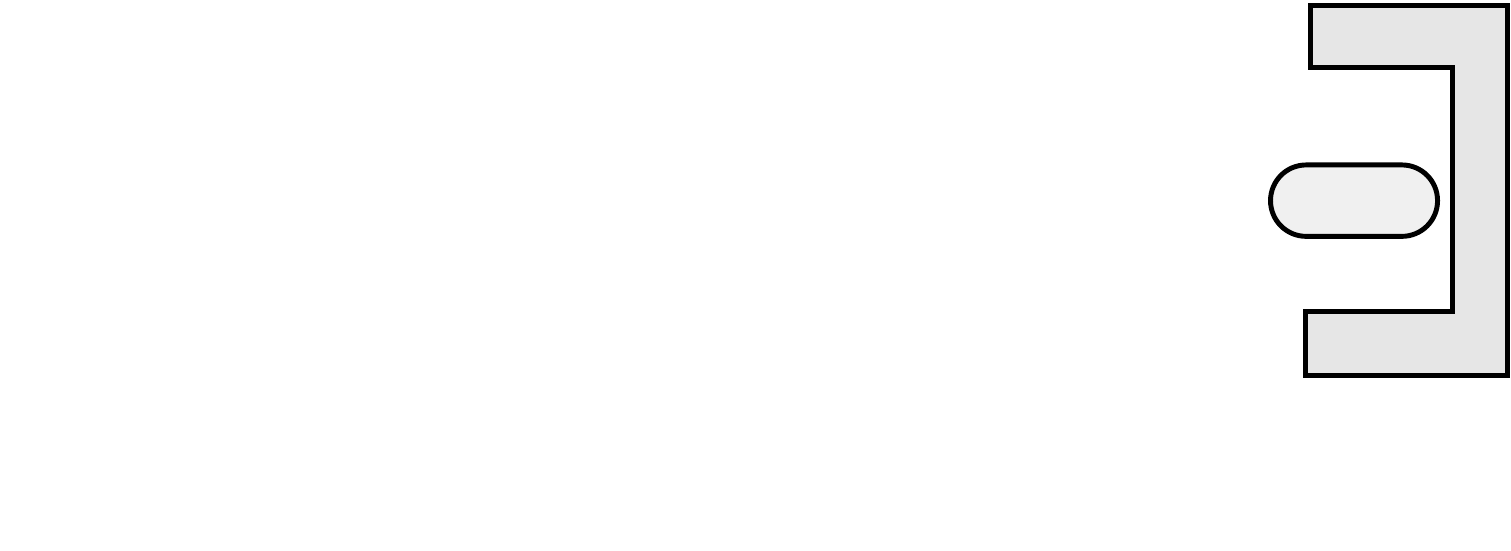
    \caption{Illustration of the most common optical sensing working principles: (a) \acrfull{ORLI}, (b) \acrfull{OTOF}, (c) \acrfull{OTRI}, and (d) \acrfull{OBB}.}
    \label{fig:optical_proximity_sensor_type}
\end{figure*}


In this section, we describe the principles, research background, and the latest research in optical sensing technology.  Optical sensing is one of the most popular and traditional forms of proximity sensing in robotics. The main principles, as shown in Fig.~\ref{fig:optical_proximity_sensor_type}, are:
\begin{itemize}
    \item \Gls{ORLI},
    \item \Acrfull{OTOF},
    \item \Gls{OTRI} and
    \item \Gls{OBB}.
\end{itemize} 
In the cases of \gls{ORLI}, \gls{OTOF}, and \gls{OTRI}, a light emitter and a receiver are placed next to each other on the same surface. Then, the proximity of an object is measured based on the reflected light intensity, return time of reflected light, or light incident position (or angle) respectively. Especially for \gls{ORLI}, a paired set of an \gls{IR} LED and a photodiode can be called \emph{photoreflector}. Furthermore, \gls{ORLI} often uses modulated light to suppress extraneous light influences. In the case of \gls{OBB}, a light emitter and a detector are arranged on distinct surfaces to detect the interruption of the ray due to an obstacle. 

With \gls{ORLI}, the proximity value depends on the reflectance of the object, which affects the measured light intensity. With \gls{OTOF} and \gls{OTRI}, the actual distance without direct dependency on the reflectance is measured. However, most instances of \gls{OTOF} and \gls{OTRI} have difficulties with specular reflections as they can occur for instance on metallic surfaces. When reflectance properties of surfaces are problematic, \gls{OBB} is an interesting alternative, as it can detect objects, even with very shiny surfaces. In robotics, the \gls{ORLI}-type has been widely used, as the sensor structure and processing are simple, easily complying with the requirements on integration. Therefore, it has been a popular choice for equipping manipulators (\gls{ATI}) and grippers (\gls{ATII}) with proximity sensors.

\subsubsection{Reflected Light Intensity-Type Sensors (\gls{ORLI}) at an Early Stage}
\label{subsubsec:RLIEarly}

In 1973, Lewis et al.~\cite{Lewis1973PlanningCF} at NASA's Jet Propulsion Laboratory (JPL) proposed a gripper with the \gls{ORLI}-type in a JPL program (see also~\cite{johnson_optical_1973}). The main goal of the program was
\begin{quote}
``[...] to demonstrate the integration of sensory and motor functions in the autonomous performance of manipulation and locomotion tasks in response to global commands issued by an operator.''~\cite{Lewis1973PlanningCF}
\end{quote}
\gls{ORLI}-type sensors are suitable for both application types described in Sec.~\ref{subsec:ApplicationsAndBehaviorTypes}, because of their small sizes and fast response times. However, in the 1970s, the sizes of LEDs, detectors, lenses, and amplifier circuits still were too large. Also, CPU performance was not sufficiently high. For this reason,  it was technically difficult to mount an array of multiple optical sensors on manipulator links or grippers. 

\begin{figure}
    \centering
    \includegraphics[width=0.45\linewidth]{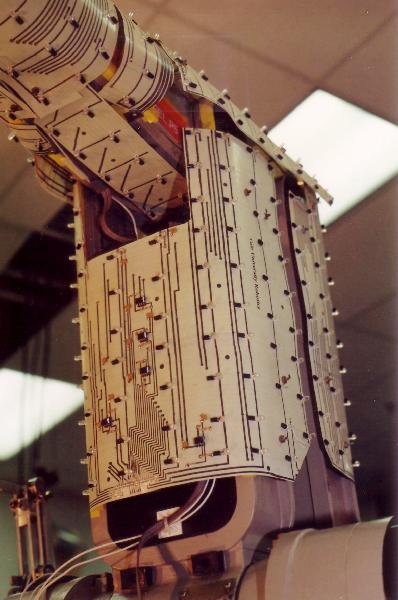}
    \caption{Using flexible circuit boards, Lumelsky and Cheung show the realization of a proximity skin covering whole manipulator arm as discussed for instance in~\cite{Cheung1992}.Sensor type: \gls{ORLI} (Photo courtesy of www.edcheung.com)}
    \label{fig:SkinCheungAndLumelsky}
\end{figure}

In the late 1980s and early 1990s, advancement in this technology already allowed Cheung and Lumelsky to show designs for an \gls{ORLI}-type proximity skin for collision avoidance tasks~\cite{Cheung1992} based on an Opto Diode Corp.\ OD8810 infra-red emitter and an Osram SFH205 photodiode as shown in Fig.~\ref{fig:SkinCheungAndLumelsky} (see also Sec.~\ref{subsec:CollisionAvoidance}). This technology was adapted by Wegerif et al.\ in \cite{wegerif1992sensor} for their own work on collision avoidance. In both cases, these solutions required the design of an analog front-end to drive the sensors and custom made electronics for the skin as a whole. Modulation of light is used to handle potential cross-talk between different sections of the skin. Similar technology is featured in the work of Petryk and Buehler~\cite{Petryk1996,Petryk1997}, who equipped a two-jaw gripper with distributed sensing. 

\subsubsection{Reflected Light Intensity-Type Sensors (\gls{ORLI}) Using Photoreflectors and Custom Electronics}

\begin{figure}
    \centering
    \includegraphics[width=\linewidth]{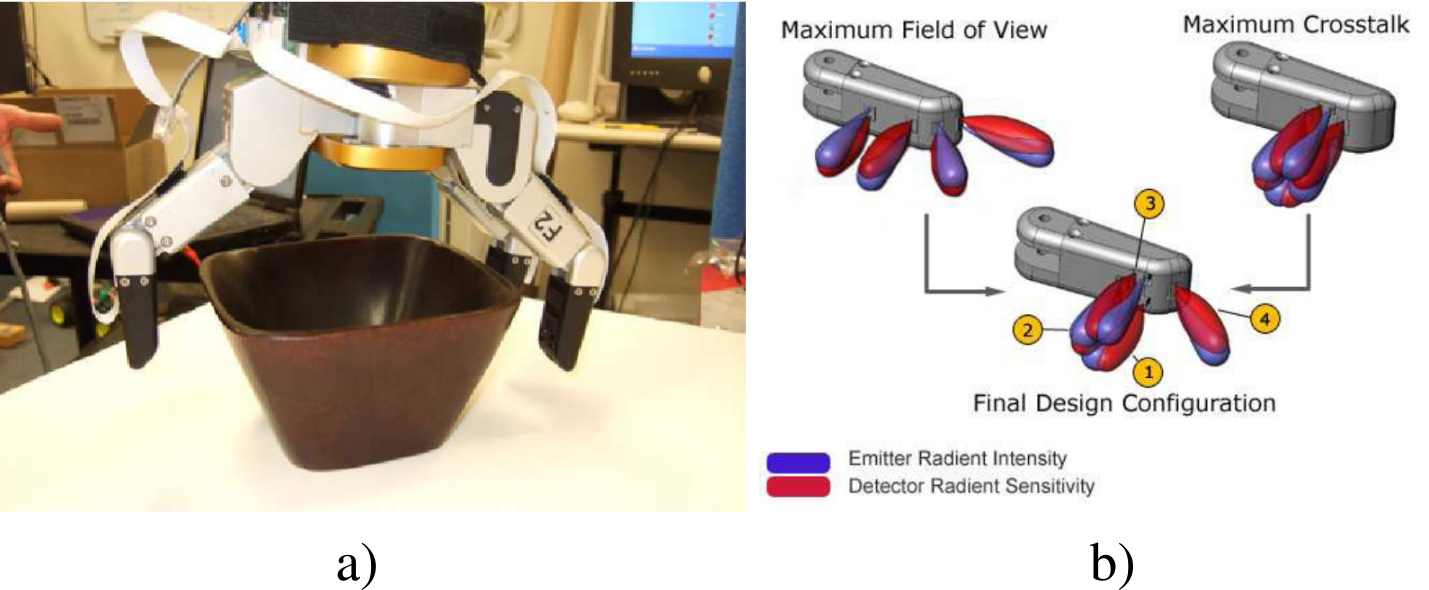}
    \caption{a) Barrett hand equipped with finger-size optical proximity sensors
    b) Design criteria leading to a final concept of sensor integration in the fingertips of the Barrett hand. ~\cite{Hsiao2009} Sensor type: \gls{ORLI} (Figures reprinted with kind permission of the authors \copyright 2009 IEEE)}
    \label{fig:Hsiao2009_final}
\end{figure}

In the 2000s, many companies released surface-mounted photoreflectors and small microcontroller/amplifier circuits. As a result, researchers were able to develop an array of multiple sensors suitable for ATs I and II more easily. An interesting example is due to Tar et al.~\cite{Tar2009}, who show the realization of an $8\!\times\!8$ matrix of sensors capable of imaging approaching objects using the TCRT1000 photoreflector. Hsiao et al.~\cite{Hsiao2009} developed an \gls{ORLI}-type sensor for the finger of a Barrett Hand as shown in Fig.~\ref{fig:Hsiao2009_final}. The sensor was constructed using four photoreflectors and an amplifier circuit/microcontroller, embedded in each finger segment. In~\cite{Mittendorfer2011}, Mittendorfer and Cheng first showed their multi-modal and modular sensor design that uses the Sharp GP2S60, having a footprint of $3\!\times\SI{4}{\square\milli\meter}$, as a proximity sensor (see Fig.~\ref{fig:LargeScaleModularSkin} as well as Sec.~\ref{subsec:Multi-ModalSensors}). A special case of \gls{ORLI}-type sensors can be implemented using optical fibers. Since the fibers are easy to integrate into confined spaces, this solution has been proposed by Espiau and Catros~\cite{espiau1980}, Walker et al.~\cite{Walker2007}, and Konstantinova et al.\
\cite{konstantinova2015force,konstantinova2016fingertip,palermo2020automatic}. In~\cite{Walker2007}, the authors integrate 32 fibers, having a diameter of $\SI{1}{\milli\meter}$, into a disc-like end-effector. The fibers route the reflected light captured around $\SI{360}{\degree}$ to a $4\!\times\!8$-display. The intensity values on the display are then recorded by a camera. In \cite{konstantinova2015force,konstantinova2016fingertip,palermo2020automatic} optical fibers are similarly routed from the tip of a finger to a signal processing module (KEYENCE) for \gls{ORLI}-type sensing.

\label{subsubsec:RLIAnalog}

\begin{figure}
    \centering
    \includegraphics[width=0.8\linewidth]{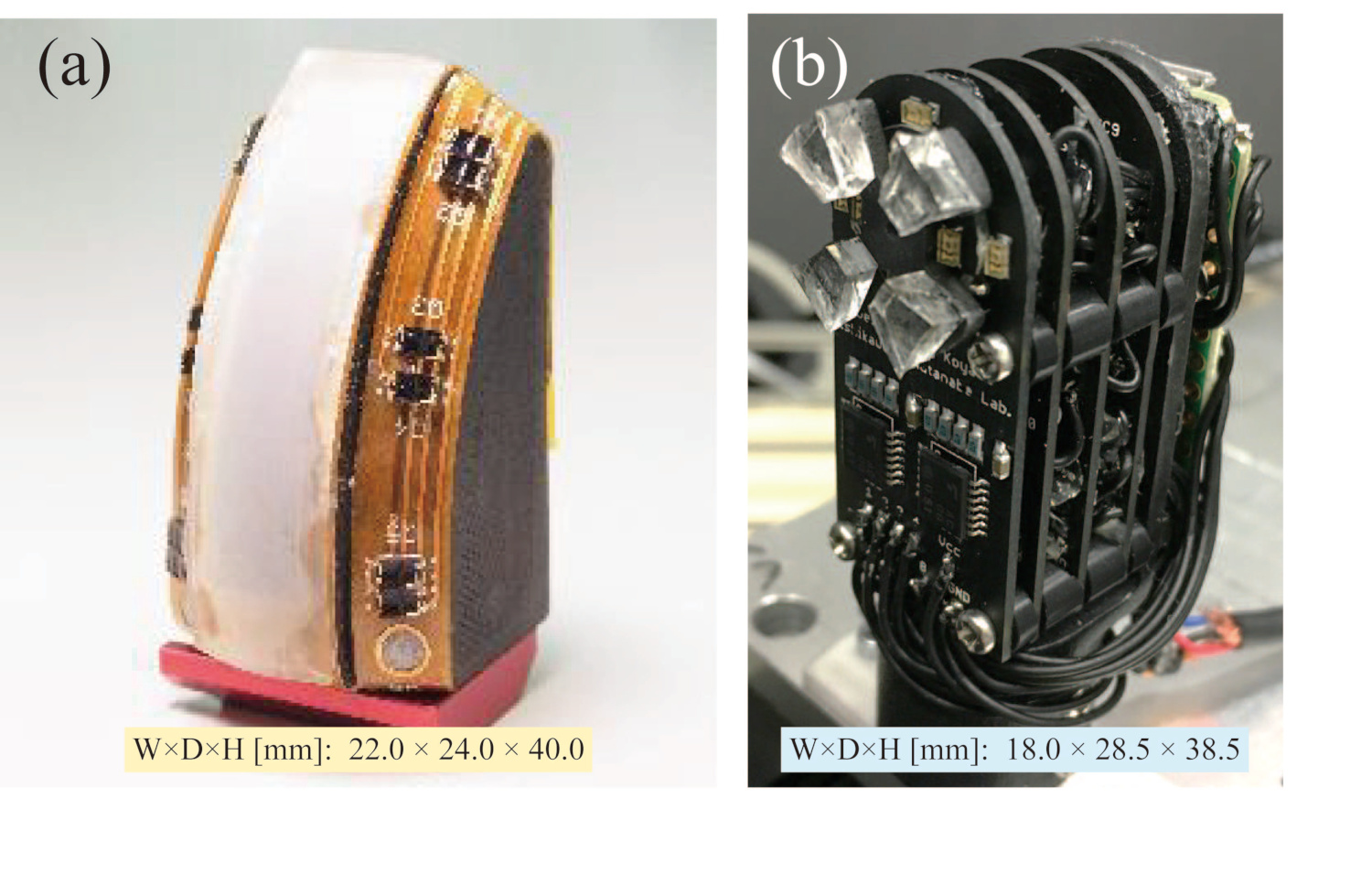}
    \caption{(a) High-speed proximity sensor~\cite{koyama2013,koyama2016,koyama_ijrr2019}.
    (b) High-speed, high-precision proximity sensor~\cite{koyama2018,koyama2019}. 
    Both sensors detect tilt angles and distance of an object's surface.
    The measurement time of both sensors is less than $\SI{1}{\milli\second}$. 
    The distance resolution of the sensor (b) is less than $\SI{51}{\micro\meter}$. Sensor type: \gls{ORLI} (\copyright 2013 and 2019 IEEE) 
    }
    \label{fig:koyama2013-2019_optical}
\end{figure}

In~\cite{koyama2013,koyama2016,koyama_ijrr2019}, Koyama et al.\ developed finger-sized, high-speed proximity sensors mounting twelve photoreflectors on a fingertip as shown in Fig.~\ref{fig:koyama2013-2019_optical}~(a). The sampling time of the sensor outputs is  $<\SI{1}{\milli\second}$, and the sensor size is thin and compact~\cite{Hasegawa2015}. As the outputs of the photoreflectors were processed in a grid of resistors, a proximity event can be localized. The same principle is adopted by Arita and Suzuki in~\cite{arita2021contact} for a linear array of sensors. The authors also proposed a simple calibration method using changes in fingertip positioning and reflected light intensity. However, simple calibration methods, e.\,g.~\cite{Hsiao2009,koyama_ijrr2019}, have relatively large errors (with millimeter or sub-millimeter accuracy) due to fingertip position errors or circuit noise. Therefore, reactive preshaping methods (see Sec.~\ref{subsec:PreshapingAndGrasping}) have not yet reached a high level of accuracy using these calibration schemes. 

\subsubsection{Reflected Light Intensity-Type Sensors (\gls{ORLI}) Using Sensing Modules)}
\label{subsubsec:RLIDigital}

More recently, some companies have released compact-sized, low-noise proximity sensors with built-in amplifier circuits and $I^2C$ bus connectivity. Multiple sensors can be daisy-chained using the $I^2C$ bus. Researchers can develop fingertip-size proximity sensors and robot skins equipped with multiple ranging sensors. In particular, Vishay Semiconductors released the \gls{ORLI}-type sensor, the VCNL4010. The footprint of the VCNL4010 is $3.95\!\times\!\SI{3.95}{\square\milli\meter}$, and can measure a range of $1\text{\textendash}\SI{200}{\milli\meter}$ within $\SI{4}{\milli\second}$ (minimum time setting). Although the sensor output is affected by the reflectance of object surfaces, the development of a thin proximity sensor is easily attainable.

\begin{figure}
    \centering
    \includegraphics[width=1.0\linewidth]{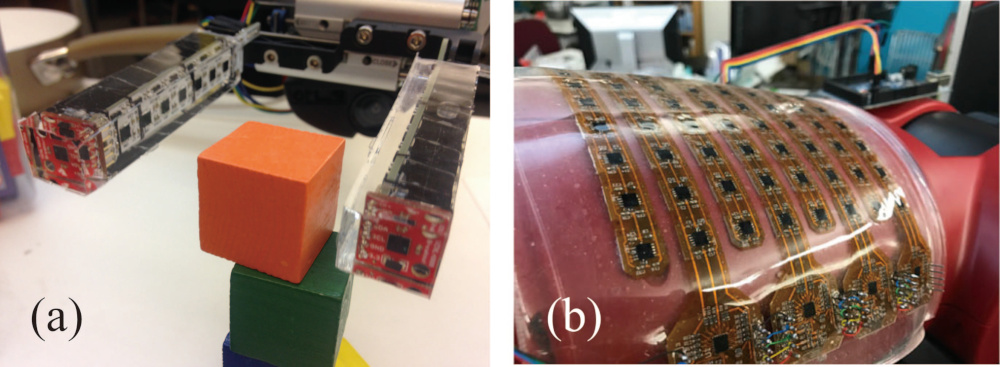}
    \caption{(a) Gripper-type proximity sensor that can detect contact, force, and distance, (b) Robot skin type. Sensor type: \gls{ORLI} (Figures reprinted with kind permission of the authors (a) \cite{patel_integrated_2018} \copyright 2018 Springer and (b) \cite{hughes_robotic_2018} \copyright 2018 IEEE)}
    \label{fig:correll2018_optical}
\end{figure}

Patel et al.~\cite{patel_integrated_2018} have developed a finger-size sensor (Fig.~\ref{fig:correll2018_optical}~(a)) that can detect distance, contact, and force with the VCNL4010. The sensor consists of multiple VCNL4010 devices covered with a transparent rubber (PDMS silicone). When there is no contact between the sensor and an object, the sensor can measure distance based on the reflected light intensity from an object's surface. The sensor can also detect contact with an object triggered by a sharp change of reflected light intensity. After contact, the contact force can be also estimated by measuring the reflected light from the object surface to the rubber surface. Hughes et al.~\cite{hughes_robotic_2018} have proposed flexible robot skin modules (Fig.~\ref{fig:correll2018_optical}~(b)) using the same sensor structure as in \cite{patel_integrated_2018}. They realized gesture recognition by combining distance values with a random forest classifier. Originally conceived for oxymetry, the MAX30105 by Maxim Integrated is used for implementing wireless multi-modal sensor for the hand of a humanoid robot (Robonaut 2) in \cite{markvickawireless}.

In \cite{Maldonado2012}, the authors repurpose a mouse sensor (ADNS-9500) in a fingertip for proximity sensing, as \gls{ORLI} is used in the sensing element. In this case, it is even possible to use the $30\!\times\!30$-pixel delivered by the sensor for further processing (e.\,g.\ texture recognition and slip detection).

\subsubsection{Time-of-Flight-Type (\gls{OTOF}) Sensors}
\label{subsubsec:ToF}

One popular \gls{OTOF} sensor is the VL6180X proximity sensor, released by STMicroelectronics. The VL6180X can measure $10\text{\textendash}\SI{100}{\milli\meter}$ with $\SI{1}{\milli\meter}$ resolution and a repeat measurement error of $\pm 1\text{\textendash}\SI{2}{\milli\meter}$. The measurement time for one sensor is $\SI{7}{\milli\second}$ to several tens of $\si{\milli\second}$, depending on the settings. Lancaster et al.~\cite{yang_pre-touch_2017,huang2018visionless} developed a fingertip-sized sensor for a parallel jaw gripper on the PR2 and demonstrated a robust manipulation of a Rubik's Cube. They also developed a fingertip-sized sensor comprised of transparent rubber and a \gls{OTOF} sensor, which uses reflected light intensity for force measurement~\cite{Lancaster2019}. They designed and evaluated different rubber shapes and optical configurations (flat rubber, rounded rubber, and light blocker configurations). It is reported that a rounded configuration improves the sensitivity of force detection. Sasaki et al.~\cite{sasaki2018robotic} developed a multi-modal proximity sensor, employing both \gls{ORLI} and \gls{OTOF} sensing. The \gls{ORLI}-type detects distance and posture for an object on a table, and \gls{OTOF} type measures the distance from the table surface. The robot can adjust the configurations of the fingertips and the end-effector (the hand base) simultaneously using sensor feedback. Tsuji et al.~\cite{tsuji2019} developed a proximity sensor skin using \gls{OTOF} sensors for a collaborative robot, which in its layout is comparable to the work by Cheung and Lumelsky (see Fig.~\ref{fig:SkinCheungAndLumelsky}). In \cite{ding2018capacitive,ding2019with}, Ding et al.\ show their developments of a multi-modal proximity sensor, the \emph{proximity sensing cuffs}, featuring capacitive and \gls{OTOF} technology for material recognition and collision avoidance.

Recently, the use of \gls{OTOF} has also been proposed for Soft Robotics devices \cite{hellebrekers2018liquid,yin2020closing}. Even though the module used is the already mentioned VL6180X, which is not deformable, the authors show its integration in a soft circuit, featuring traces of copper wetted with eutectic gallium indium (EGaIn) inside a thin PDMS sheet. The circuit features other sensors ICs: an IMU, barometric pressure, and temperature sensors. The soft sheet is then used to equip a two-jaw gripper with these sensing capabilities. 

\subsubsection{Triangulation-Type Sensors (\gls{OTRI})}

An early example for an \gls{OTRI}-Type proximity sensor is due to Fuhrman and Kanade~\cite{fuhrman1984}. Using a chip capable of localizing a light spot, they realize several light sources that are evaluated in a time-multiplexed manner, resulting in an object's proximity value as well as orientation and curvature. However, this setup can probably be considered to be too bulky, i.\,e.\ not skinlike, for modern applications. In \cite{ceriani2013optimal,avanzini2014safety}, the Ceriani et al.\ and Avanzini et al.\ report using the Sharp GP2Y0A02YK0F module, that guarantees a consistent distance output across different reflectivity of surfaces. In their work, they explore the optimal distribution of sensing elements for safe HRI (see also Sec.~\ref{subsubsec:RecentJacobian}). To measure distance and posture more precisely, Koyama et al.~\cite{koyama2018,koyama2019} developed a high-speed, high-precision proximity sensor, as shown in Fig.~\ref{fig:koyama2013-2019_optical}~(b). The sensor has the size of a human fingertip ($18\!\times\!28.5\!\times\!\SI{38.5}{\cubic\milli\meter}$), and it can detect the distance to and postures of an object surface with a distance error of fewer than $\SI{31}{\micro\meter}$ and a measuring time less than $\SI{1}{\milli\second}$. A similar sensor design was explored by Bonen et al.~\cite{bonen1997}, who propose a single emitter and multi-detector architecture for detecting distance as well as object orientation.

\subsubsection{Break-Beam-Type Sensors (\gls{OBB})}
Teichmann et al.~\cite{Teichmann2000} mounted a light-emitting diode at one end of a parallel jaw gripper and detectors at the other end, and switched to reactive motions based on light blocking due to an object. They also describe the application of this approach to a three-fingered hand. In \cite{Guo2015} Guo et al.\ show the integration of an array of \gls{OBB}-Type sensors in the jaws of PR2's parallel gripper for reactive preshaping and grasping challenging objects, where other approaches would fail, such as semi-transparent tissues. 

\subsubsection{Discussion on Optical Sensors}

The devices implementing optical sensing are small in size and have a high-speed response. These advantages are suitable for sensor/actuator integration and automatic grasping using a robot hand, although the sensing has difficulty detecting transparent, black, and shiny objects. To detect all these objects, it is necessary to introduce multi-modal sensing, such as a combination of optical and capacitance sensing.

In terms of spatial resolution, optical sensing elements can be quite small. For example, the Sharp GP2S60 photoreflector used by Mittendorfer et al.~\cite{Mittendorfer2011} has a footprint of $3\!\times\!\SI{4}{\square\milli\meter}$, potentially allowing a density of a few elements per $\si{\square\centi\meter}$. However, on large-area skin (\gls{ATI}) such high densities can be impractical in terms of the electronic effort needed (wiring, signaling, etc.). Regarding sensing range, \gls{ORLI}-types have been reported to produce a relatively large detection distance of about $\SI{300}{\milli\meter}$~\cite{Cheung1992} (\gls{ATI}), but these sensors are also often used for lower ranges, e.\,g.~\cite{koyama2018}, where a maximum detection range of $\SI{20}{\milli\meter}$ is reported (\gls{ATII}). In both cases, there is a small dead spot near the sensing element. Typical \gls{OTOF}-technology used, e.\,g.~\cite{ding2019with,tsuji2020proximity}, can work up to a distance of $\SI{4}{\meter}$, depending on the component used, but this extended range has the cost of having a relatively large dead spot of $\SI{10}{\centi\meter}$ in front of the sensing element. Similarly, the \gls{OTRI}-Type sensor used in~\cite{ceriani2013optimal,avanzini2014safety} has a detection range of $\SI{1.5}{\meter}$ and a dead spot of $\SI{20}{\centi\meter}$. 

\subsection{Radar}
\label{subsec:Radar}

In recent years, radar sensing technology has become popular in human-centered technologies due to the development of system on chip radar systems reducing the size, which also makes them very attractive for integration on robotic platforms. Recent developments are driven in big part by the automotive industry (see also~\cite{zhou2020mmw}). Radar sensors withstand harsh weather environments and can augment widely used optical sensor technologies, meaning they have crucial traits for enabling highly automated driving. An important aspect for radar sensors in human-robot interaction is that it is a technology that is widely used in safety related applications in the automotive domain. Consequently, existing expertise from the automotive domain can potentially be utilized towards robotic applications (e.\,g.~\cite{Gerstmair_2019}). A recent survey by van Berlo et al.\ summarizes the current application fields of radar technologies~\cite{van2020millimeter}, including the aforementioned domain of automotive and \gls{HCI} (tracking, gesture recognition, etc.).

\begin{figure*}
    \centering
    \fontsize{8}{10}\selectfont
    \includegraphics[width=0.75\textwidth]{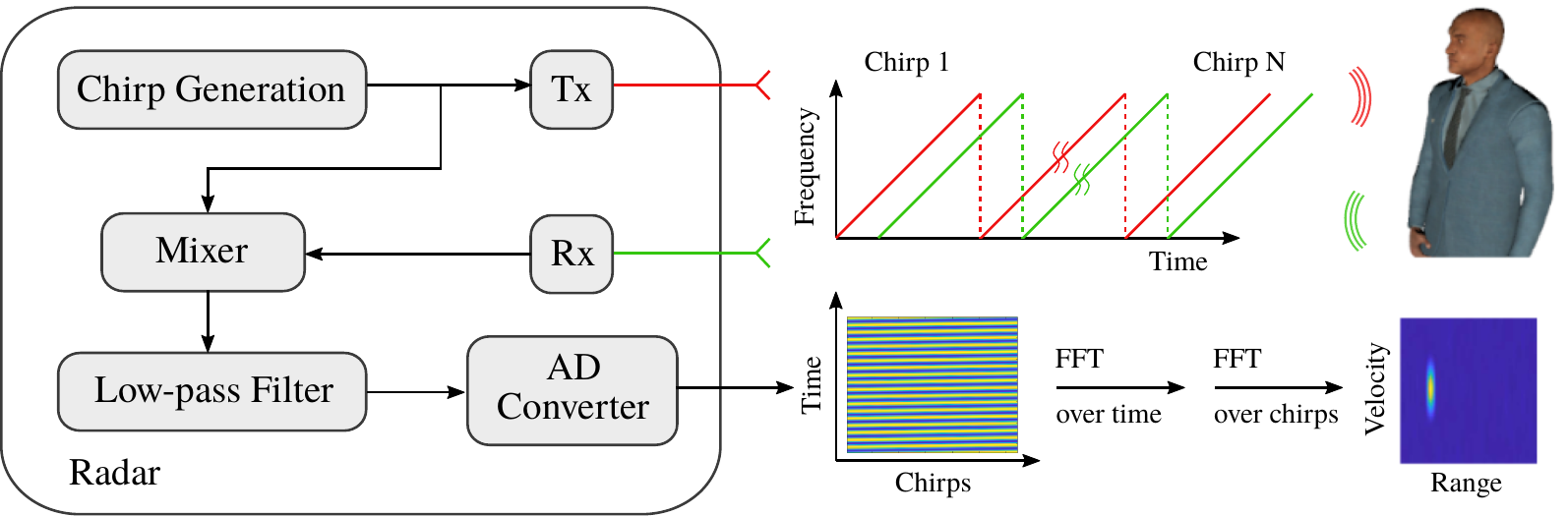}
    \caption{Principle of FMCW Radar: A transmitter (Tx) sends out a chirp signal (red), which gets reflected at object boundaries, e.\,g. a human. The reflected signal (green) at the receiver (Rx) is a delayed, attenuated copy of transmitter signal. The time delay corresponds to the distance and thus the frequency difference between transmitter and receiver is proportional to the distance. This frequency is obtained by mixing the transmitter and receiver signal. For a sequence of chirps, the phases of the received signal changes due to the Doppler shift. An FFT over the time extracts the frequencies, a subsequent FFT over chirps extracts the velocity, such that a 2D range distance map is obtained.}
    \label{fig:radar}
\end{figure*}

Recently, in a joint effort, Google and the chip manufacturer Infineon boosted this technology as they introduced a $\SI{60}{\giga\hertz}$ radar chip with integrated transmitter and receiver antennas for fine gesture interaction based on \gls{FMCW}~\cite{Lien2016}. The principle of FMCW radars is illustrated in Figure~\ref{fig:radar}. A more detailed description can be found e.\,g. in \cite{Gerstmair_2019}.

Advantages of \gls{FMCW}-radar are that they provide distance and velocity measurements simultaneously with a high resolution for close ranges, which makes them suitable for proximity perception, collision avoidance, and \gls{HRI}, e.\,g.\ gesture control, in the field of robotics. In~\cite{stetco2020}, a simulation approach for \gls{FMCW}-radar is proposed and evaluated for collaborative robotics. The simulation approach aims at fast development and evaluation of signal processing algorithms as well as machine learning approaches for \gls{FMCW}-radar before deploying it on a robot (sim-to-real learning). Multi-modal tactile and radar sensing for grasping applications and object classification was introduced in~\cite{Flintoff2018}.

Further manufacturers, such as SiliconRadar, Texas Instruments, etc.\ also provide system on chip radar solutions. In~\cite{Geiger2019}, the integration of a multiple antenna radar sensor based on $\SI{160}{\giga\hertz}$ with flexible waveguides for collaborative robots was studied. Another operation mode for radar technology is \gls{SAR}. In robotics, it is utilized for instance for mobile robot navigation. In~\cite{watts2016} \gls{SAR} based radar imaging was investigated on a PR2 robot system for 2D and 3D imaging of objects in close proximity to show the potential of radar sensing to complement optical sensors in robot perception and manipulation.

Besides proximity and distance sensing, radar systems can be applied to exploring and constructing subsurface 3D maps \cite{gssi2021gpr}. Those radars are known as \gls{GPR} and are applied in different fields such as assessment of dense underground utilities in urban areas, evaluation of the subsurface for energy and mineral production operations or the detection of buried objects, on earth or potentially on other planets. Its sensing principle is based on emitting radar pulses and evaluating the propagation velocity into the subsurface and the soil utilities \cite{kouros20183d}. 

\subsection{Other Sensing Principles}
In this section, we introduce further sensing principles that can be used for proximity sensing, i.\,e.\ acoustic, inductive, and whiskers. As of 2021, they can still be considered to be less \emph{mainstream} in human-centered robotics than the capacitive, optical, or radar ones. However, they offer interesting alternatives and can outperform other principles discussed so far in some scenarios.

\subsubsection{Acoustic}

The widest spread technique for ranging based on acoustic wave propagation is \gls{AUS}, which can be found in many domains, particularly for under-water ranging. In robotics, this technology is easily available for the enthusiast and professional use, such as the MaxSonar-series by MaxBotix. 
Higher-end solutions, featuring 3D echolocation are also available, for instance by Toposens. 
Nunes et al.\ proposed the use of ultrasound in \cite{Nunes1994} for 3D ranging in 1994. Even though Dario et al.~\cite{Dario1990} propose an ultrasound sensor for integration into a fingertip, the sensors usually have a non-negligible offset or dead-spot for sensing around the sensing element. Therefore, many of the available solutions are non-practical for pre-touch applications, i.\,e.\ close proximity (see Fig.~\ref{fig:SensingRange}). Integration is also challenging because sensing elements do not scale down easily. Thus, oftentimes use-cases of \gls{AUS} are more similar to laser-range finders, i.\,e.\ mid-range and long-range sensing. In~\cite{fang2019toward}, Fang et al.\ circumvent the mentioned difficulties by mounting the sensor at a distance and tilted while bouncing the waves off a parabolic mirror. However, the integration remains limited to one acoustic sensing element. Ultrasound is also widely used in parking sensors. An example of a combination with capacitive sensors to overcome detection limitations at short distances is provided in \cite{Schlegl2011}. The group of Prof.\ Steckel at University Antwerp has a strong focus on 3D \gls{AUS} for robotic applications, e.\,g.~\cite{steckel2017acoustic,verellen2020high}. The group has achieved remarkable results in areas such as SLAM for the navigation of mobile platforms. However, as with other designs, the ultrasound sensing platform is not very \emph{skinlike}, thus limiting the pre-touch applications in favor of longer-range sensing. Furthermore, the authors often make a point to establish this technology as an alternative to LIDAR and other mid-range or long-range sensing options for ground vehicles but also for \gls{UAVs}. The use of \gls{AUS} in air-borne vehicles puts in evidence that this type of sensing can be considered to be bio-inspired by bats and their echolocation capabilities.

Another type of acoustic sensing has been proposed by Jiang et al.~\cite{Jiang2012,Jiang2013}, which the authors call the \emph{\gls{AS}}. A microphone is placed inside a cavity that is worked into the structure of a finger. As a surface approaches the opening, the resonance frequency of the cavity changes. By analyzing the differences in the spectrum between an external microphone and the microphone inside the cavity, the distance can be estimated. This works for very close range (up to $\approx\SI{4}{\milli\meter}$). In their work, this modality is explored, because it does not suffer from detection difficulties related to transparency or reflections (optical sensing) or low dielectric contrast (capacitive sensing).

\subsubsection{Inductive}

Inductive sensors utilize alternating magnetic fields to detect objects, as they disturb the generated magnetic field, which can be detected as a change of inductance of a coil, a change of the mutual inductance between several coils or directly by measuring the magnetic field. The objects do not need to be ferromagnetic. In particular, objects with high conductivity such as metals, strongly affect an alternating magnetic field as eddy currents near the surface of such objects prevent deep penetration of the materials by the magnetic fields. Inductive proximity sensors are very robust and commercial sensors provided by a variety of manufactures are widely used in industry as proximity switches, typically detecting conductive objects. However, as these commercial or industrial sensors are not found in robotics, they are not included in this survey. The capabilities of inductive sensors for non-metallic objects are more limited and inductive sensors have therefore been used in combination with other approaches to classify materials, for example.

The sensing system proposed in \cite{han2016highly} is stated to combine capacitive force and inductive proximity sensing with a range of up to $\SI{150}{\milli\meter}$ for conductive materials with the help of a layer of \gls{CMC}. The sensor was then enhanced and in \cite{nguyen2017highly}, having a higher detection range and spatial resolution. The \gls{CMC} layer was used to form an LCR circuit and enable both tactile and proximity sensing. In a related work \cite{nguyen2020skin}, an electromagnetic field was formed by exciting a combined co-planar plate capacitor and a coil embedded in a flexible circuit board. The impedance of the resulting LCR circuit was analyzed and the relationship to the distance of different objects was presented as proximity measurement, with a sensing range of up to $\SI{300}{\milli\meter}$.

In \cite{George2010} the combination of capacitive and inductive sensing is used to distinguish between humans and other objects such as (grounded) laptops. While the capacitive signals for humans and grounded laptops are very similar, the inductive signal is much different, as the laptop comprises highly conductive metallic parts and thus has a stronger influence on the magnetic field. Even though the setup is not intended for proximity sensing, a range of up to $\SI{150}{\milli\meter}$ is reported. Consequently, the system can also be used to detect non-conductive and conductive objects and offers a very high measurement rate of $\SI{25}{\kilo\hertz}$. With multiple coils, inductive sensors can not only be used to obtain a distance estimate but full 6 DoF information, as discussed in \cite{Gietler2019}.

\subsubsection{Whiskers}
Finally, on the fringe of the domain of proximity perception, we can find \emph{artificial whiskers} that are inspired by mammals, such as rodents, who use them to navigate and explore their environment \cite{russell2003object}, \cite{lepora2018tacwhiskers}. These whiskers are beams that bend due to external forces (contacts with walls, wind, etc.) and usually, the resulting force/deformation at the base is measured. These approaches are often featured as part of the tactile perception community, as sensing is actually contact-based, but they are used to probe the nearby environment much in the same way a proximity sensor is used.

\subsection{Multi-modal and Modular Sensors}
\label{subsec:Multi-ModalSensors}

The possibility of deploying proximity sensors alongside other sensing modalities on robots (vision, touch, etc.) is a key aspect of the success of this technology. Only if they coexist with the other modalities, can they fill the perception gap that is left by them. This challenge is evidenced by the many existing realizations of multi-modal sensors, especially by designs that include the tactile modality alongside the proximity one. Furthermore, it is common to find that these designs are conceived in a modular manner. The \emph{HEX-o-Skin} by Mittendorfer et al.~\cite{Mittendorfer2011} is a prominent example of this trend. It is a modular design, which is suited for covering large areas of the robot (see Fig.~\ref{fig:LargeScaleModularSkin}) and includes proximity, tactile, inertial, and temperature sensing in each unit. 

The most generic approach for implementing multi-modal sensing is to use a specialized measurement principle for each desired modality. The work by Mittendorfer et al.~\cite{Mittendorfer2011}, again, is an example of this approach. Proximity detection is implemented by \gls{ORLI} and tactile events are detected with a capacitive sensing element. Stiehl et al.~\cite{Stiehl2006}, who implement capacitive proximity (\gls{CSE}, provided by the MC33794), force and temperature measurement in a pseudo-modular skin, which the authors argue helps in distinguishing social contacts from collisions with the environment. Another, less modular example is due to Guan et al.~\cite{Guan2012}. In their work, they equip the gripper of a climbing robot with a range finder sensor, two ultrasound modules, and a camera. 

In~\cite{han2016highly}, Han et al.\ show a tactile proximity sensor where the proximity modality principle is inductive and the tactile modality is capacitive. In~\cite{konstantinova2015force,konstantinova2016fingertip,palermo2020automatic} proximity sensing and tactile (force) sensing is implemented using optical fibers. In proximity sensing, the \gls{ORLI}-type is used, for tactile sensing, the reflection that changes inside a movable part is measured. As explained in Sec.~\ref{subsubsec:RLIDigital}, in~\cite{patel_integrated_2018} the authors show the implementation of a tactile proximity sensor based on \gls{ORLI}-type sensing alone. As stated before in Sec.~\ref{sec:CapacitiveSensing}, capacitive sensing is especially attractive for joint tactile and proximity designs. Example designs are shown in \cite{Lee2009,Goeger2010a,Goeger2013,Alagi2016a}. Designs also have been proposed outside of robotics literature, which are nonetheless potentially relevant, e.\,g.~\cite{Zhang2014}. Finally, there is a subset of approaches that utilize different measurement principles for redundant proximity sensing. This is the case with Ding et al.~\cite{ding2018capacitive,ding2019with}, and Tsuji et al.~\cite{tsuji2020proximity} that use both an \gls{OTOF} and \gls{CSE} for robustness. Markvicka et al.\ propose the joint use of \gls{OTOF} and \gls{ORLI} in~\cite{markvickawireless}.

\section{Applications and Methods in the Research and Industry Domains}
\label{sec:Applications}

In this section, we will review the contributions from the field focusing on the applications and methods presented. We will follow the organization presented in Sec.~\ref{subsec:ApplicationsAndBehaviorTypes} and Fig.~\ref{fig:ApplicationsAndBehaviors}, i.\,e.\ focusing separately on \gls{ATI}  
and \gls{ATII} and going from low-complexity behaviors (\gls{BTI}) to high-complexity behaviors (\gls{BTII}).
\subsection{Reactive collision-avoidance and Contour Following (\gls{ATI}, \gls{BTI})}
\label{subsec:CollisionAvoidance}

Collision-avoidance is regarded as a fundamental skill for autonomous robots as well as for safe human-robot interaction. To start this section, we want to motivate with a quote by Novak et~al.\ from their 1992 work on whole-arm collision-avoidance, which is an elegant statement of this problem: 
\begin{quote}
``[...] since it is desirable to continue purposeful motion in the presence of obstacles, the sensor system must be able to deliver spatially-resolved proximity data, which reflects the distance to the obstacle, as well as the location along the robot and corresponding robot surface normal. This vector information may then be used to modify trajectories to permit (if possible) continued progress toward the final destination.'' \cite{Novak1992}
\end{quote}

The first important wave of interest surrounding proximity perception for collision-avoidance was sparked in the late 1980s and early 1990s~\cite{samson1990}. At that time, getting 3D information of the surroundings of the robot via cameras was challenging from a technological point of view, on the accounts of the lack of hardware and lack of performance of the CPUs. Proximity sensors, having desirable properties (low latency, skinlike), were considered an attractive alternative for this challenge. 

\subsubsection{Early Jacobian-type Approaches}
\label{subsubsec:EarlyJacobian}

\begin{figure}
    \centering
    \def\svgwidth{200pt}
    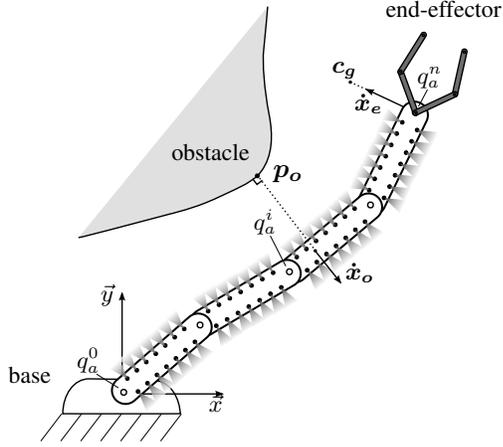
        \caption{A redundant robot, equipped with proximity sensors on its links, can move towards a target end-effector configuration while simultaneously moving some of the links of the robot away from an obstacle. (Figure adapted from~\cite{maciejewski1985obstacle})}
    \label{fig:Concept_Jacobian}
\end{figure}


A good portion of early works on collision-avoidance get inspiration from the work of Maciejewski and Klein~\cite{maciejewski1985obstacle}, published in 1985. This work introduces the notion of the ``obstacle avoidance point Jacobian'', which is analogous to the end-effector Jacobian, i.\,e.\ it relates the instantaneous joint velocity to the velocity of an obstacle point on the robot (see Fig.~\ref{fig:Concept_Jacobian}). In this figure, we use the following notation: ${q}_{a}^{i}$ are the degrees of freedom of the manipulator arm in joint-space. $\bm{x_{e}}$ and $\bm{x_{o}}$ are the end-effector configuration and the obstacle point in task-space respectively (boldface indicates these are vector quantities), whereas $\bm{c_{g}}$ is the goal configuration established for the end-effector and $\bm{p_{o}}$ is the point on the obstacle nearest to the manipulator arm (also vector quantities). Following this notation, the end-effector Jacobian $\bm{J_{e}}$ relates the velocities in the configuration space to the velocities in the task space by:
\begin{equation}
    \bm{\dot{x}_{e}} = \bm{J_{e}}\bm{\dot{q}_{a}}.
\end{equation}
Using the obstacle Jacobian $\bm{J_{o}}$, The relation in the case of the obstacle-point is likewise:
\begin{equation}
    \bm {\dot{x}_{o}} = \bm{J_{o}} \bm{\dot{q}_{a}}.
\end{equation}
Applying the same principles of using the (pseudo-) inverse of the end-effector Jacobian $\bm{J_{e}}^{+}$ for finding desired joint velocities, one can invert the obstacle point Jacobian to find the joint motions to follow a desired trajectory with respect to the obstacle point. In the case of collision-avoidance, the natural choice is a motion away from the obstacle, as indicated by $\bm{\dot{x}_{o}}$ in Fig.~\ref{fig:Concept_Jacobian}. Also, in the presence of redundancies, this approach allows projecting the avoidance motion into the null-space of a higher-order task, e.\,g.\ following a desired end-effector trajectory:
\begin{equation}
    \bm{\dot{q}_{a}}= \bm{J_{e}}^{+}\bm{\dot{x}_{e}} + (\bm{I}-\bm{J_{e}}^{+}\bm{J_{e}})\bm{J_{o}^{+}}\bm{\dot{x}_{o}}.
\end{equation}
Since $\bm{\dot{x}_{o}}$ represents the desired motion away from the obstacle in task space,  $\bm{\underline{\dot{q}_{a}}} = \bm{J_{o}^{+}}\bm{\dot{x}_{o}}$ is the joint-motion that moves the obstacle-point away from the obstacle. $(\bm{I}-\bm{J_{e}}^{+}\bm{J_{e}})$ is the expression that projects this motion into the null-space of the higher-order task, i.\,e.\ the desired end-effector trajectory. In this framework, the tasks can be ordered in a different hierarchy as well, i.\,e.\ prioritizing the collision-avoidance task over the desired end-effector trajectory.

Among the firsts to apply these ideas to proximity sensor streams were Wegerif et~ al.~\cite{wegerif1992sensor,wegerif1993whole} as well as Tamasy in \cite{tamasy1997smart} at Merrit Systems Inc. in the early 1990s. Wegerif et~al.\ studied the use of several proximity sensing technologies (IR, ultrasound, capacitive), but they ultimately covered three links of a PUMA 600 robot with a total of about 120 IR sender and receiver pairs. They modified the kinematics of the PUMA 600 to have three rotational joints in one plane, introducing a kinematic redundancy in an otherwise non-redundant robot. In their collision-avoidance algorithm, they gave the highest priority to the collision-avoidance task. They report successfully testing the system in an autonomous and a teleoperated scenario with static and dynamic obstacles. In the work by Tamasy~\cite{tamasy1997smart}, the previous work is extended by presenting the realization of a \acrlong{SNN}. These kinds of networks provide the base for equipping whole arms with proximity sensors, implementing a bus-system. IR, ultrasound and capacitive sensors can be readily connected to the system, provided they offer digitized data streams. The NASA \gls{PIPR}, featuring 18 DoFs, is shown as an application. A whole control architecture is discussed, with a GUI for user inputs, the generation of low-level commands, as well as a collision-avoidance system based on the previous developments, together with a quadratic programming approach for finding the optimal joint velocities.

Other authors that were inspired by the approach of Maciejewski and Klein are Novak and Feddema~\cite{novak1993collision,Feddema1994}. A prior work by the authors that leads up to these results is \cite{Novak1992}. In \cite{Novak1992} the authors concentrate on the development of the so-called \emph{\gls{WHAP}}-sensor, which is a skin that is comprised of mutual capacitive sensing elements for proximity sensing. In this work, the sensor is described and characterized in depth. Then, the \gls{WHAP} sensor is installed and tested on a 2-link planar robot with a total of 8 sensing elements (two per link). The robot is shown to successfully circumvent one obstacle made out of concrete and another metallic one. Later, using a sensor Jacobian in \cite{novak1993collision,Feddema1994} they concentrate on a teleoperation scenario using a 6-DoF robot arm (PUMA 560). An obstacle is responsible for a reduction of the speed of the affected DoFs as the sensors detect the obstacle approaching. However, the system is designed to not automatically move away from the obstacle, as the authors consider that this behavior is not desired by the user, at least not in a teleoperation scenario.

\subsubsection{Early Geometric Approaches}
In contrast to the Jacobian-based approaches, the collision-avoidance can be implemented by a \emph{geometric approach}, i.\,e.\ estimating features of the obstacle's surface and following its contour for as long as it obstructs the direct path from the current configuration $\bm{c_{i}}$ to a given goal configuration $\bm{c_{g}}$ of the robot. We consider this to be a geometric approach because in some way the surface of the obstacle has to be reconstructed. Early, seminal work is due to Lumelsky and Cheung \cite{cheung1988motion,cheung1989development,Cheung1992,Lumelsky1993} (and more), who propose to move along the tangent plane of the obstacle represented in configuration space. Their hardware is characterized by the use of infra-red sender and receiver pairs that are mounted on flexible printed circuit boards (which would later inspire the work by Wegerif et al.\ in \cite{wegerif1992sensor}, see above). With the flex-technology, they achieve full integration of the skin onto a manipulator as shown in Fig.~\ref{fig:SkinCheungAndLumelsky}. In~\cite{Cheung1992}, the development of the skin is explained in great detail. In~\cite{Lumelsky1993} they showcase the methods and technology in the context of teleoperation in the presence of one or more dynamic obstacles.  

\begin{figure}
    \centering
    \def\svgwidth{150pt}
    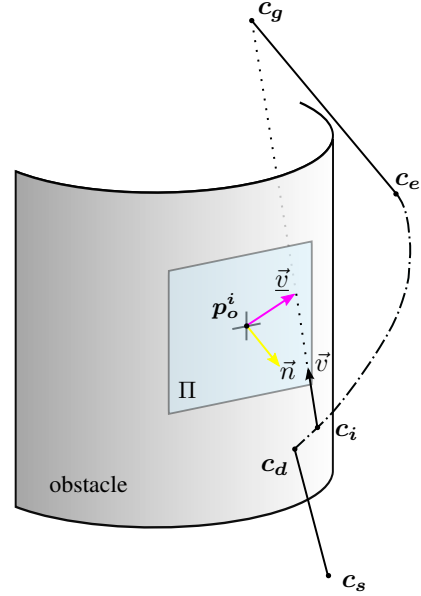
    \caption{In geometric collision-avoidance approaches, the surface of the obstacle is estimated. From this, a contour following motion can be derived.}
   \label{fig:ContourFollowingConfigurations}
\end{figure}

Nunes et al.\ \cite{Nunes1994} propose a collision-avoidance system that can be regarded as a contour following system that works at the end-effector level, i.\,e.\ in Cartesian space. As before, the idea is to slide parallel to the obstacle tangential plane. In Fig.~\ref{fig:ContourFollowingConfigurations} the geometric approaches are illustrated and summarized. It shows the four characteristic configurations $\bm{c_{s}}$, $\bm{c_{d}}$, $\bm{c_{i}}$, $\bm{c_{e}}$ and $\bm{c_{g}}$ that can be usually be identified. $\bm{c_{s}}$ is the starting configuration, $\bm{c_{d}}$ is the configuration where the obstacle is first detected, $\bm{c_{i}}$ is the current robot configuration during the contour following phase, $\bm{c_{e}}$ is the configuration where the obstacle no longer obstructs the direct path to the goal and $\bm{c_{g}}$ is the goal configuration. The vector showing from the current configuration $\bm{c_{i}}$ to $\bm{c_{g}}$ is $\vec{v}$ and its projection on the surface tangent plane $\Pi$ is $\vec{\underline{v}}$. The tangent plane is detected at a point $\bm{p_{o}^{i}}$. The dashed line represents the trajectory of the robot during the contour following procedure. In general, during contour following, the movement of the robot will be parallel to $\vec{\underline{v}}$, i.\,e.\ towards the target configuration.

\subsubsection{Recent Jacobian-type Approaches}
\label{subsubsec:RecentJacobian}

In~\cite{ceriani2013optimal} Ceriani et al.\ discuss the placement of proximity sensors on an industrial manipulator. They present an optimization method that allows to find a suitable arrangement of triangulation type sensors on the links of the robot based on the concept of a \emph{Danger Field}. The Danger Field is a distance-based metric for assessing the danger emanating from a moving robot for a human operator. Avanzini et al.\ continue this work by developing a safety control scheme on top of this concept~\cite{avanzini2014safety}. They show that tasks can be deactivated according to a predefined priority to permit evasive movements. In their case, the tasks are defined by a Cartesian trajectory split in position and orientation. Maintaining the orientation is the lower priority task, which is the first to be abandoned to comply with the collision-avoidance.

In~\cite{Schlegl2013b}, Schlegl et al.\ show the use of a capacitive sensor that features both a single-ended and a mutual-capacitive sensing capability, resulting in so-called \emph{Virtual Whiskers}. The use of both modes increases robustness in the detection of conductive and non-conductive objects. An arrangement of 7 electrodes is mounted on a segment near the wrist of a KUKA LWR 4. The hardware has a sample rate of up to $\SI{1}{\kilo\hertz}$. The authors show the combination with the on-line trajectory generation discussed in \cite{kroger2010line} that is capable of generating smooth trajectories with at least the same frequency. This leads to a highly reactive collision-avoidance prototype. Similar hardware is used in \cite{MuehlbacherKarrer2016a} M\"ulbacher-Karrer et al.\ to show the contactless control of a 9 DoF redundant manipulator. One link is equipped on two sides with electrodes. A 2D tomographic image can be extracted from each arrangement. The center of the detected event is used to steer the avoidance motion of the link equipped with sensors, while the robot continues to execute a pick-and-place task.

In~\cite{ding2019with,ding2020collision} Ding et al.\ present a Jacobian-type collision-avoidance scheme that is based on optimization, also taking into account the redundancy capabilities of the robot used. Their 7-DoF robot has three links equipped with $\SI{360}{\degree}$ sensing capability provided by so-called proximity-sensing-cuffs~\cite{ding2018capacitive}. When an obstacle is detected in~\cite{ding2019with}, joint-velocities are calculated according to a mixture of criteria, which are simultaneously optimized: distance to the target, manipulability, deviation from desired task motion, and total magnitude of joint-velocities. This results in a reflex-like collision-avoidance system, including movement parallel to the obstacle tangent plane. This can avoid getting stuck in front of obstacles, which is a possible failure mode of potential field approaches, like the one proposed by Khatib et al.~\cite{khatib1986real}. Also following up on the Jacobian-type approaches is the work presented by M'Colo et~al.\ in~\cite{mcolo2019obstacle}. The capacitive sensing technology of \emph{FOGALE robotics}~\cite{FOGALErobotics} is featured in a robotic system that has been extensively covered with electrodes to implement a skin. Using the skin, the robot can avoid static and dynamic obstacles. Finally, Arita and Suzuki show progress towards an approach that allows using force control for the seamless transition between pre-touch and touch states and for achieving desired contact forces~\cite{arita2021contact}.

\subsubsection{Recent Geometric Approaches}

For what concerns contour following scenarios, the literature proposes to include curvature estimates to improve the performance by using a predictive component, for instance by Baeten and De Schutter \cite{baeten1999improving}. Relying on the spatial resolution of the sensor, they can be estimated directly from the current sensor values, as proposed in the work by Walker et  al.~\cite{Walker2007}. This work uses an optical proximity sensor attached to an end-effector and featuring $\SI{360}{\degree}$ vision in a plane. In~\cite{Escaida2016a}, these ideas are generalized by Escaida~Navarro et al.\ for contour following in 3D, i.\,e.\ by detecting the 2D curvature of the obstacle surface.
\subsection{Reactive Preshaping and Grasping (\gls{ATII}, \gls{BTI})}
\label{subsec:PreshapingAndGrasping}

\begin{figure}
    \centering
    \footnotesize
    \def\svgwidth{240pt}
    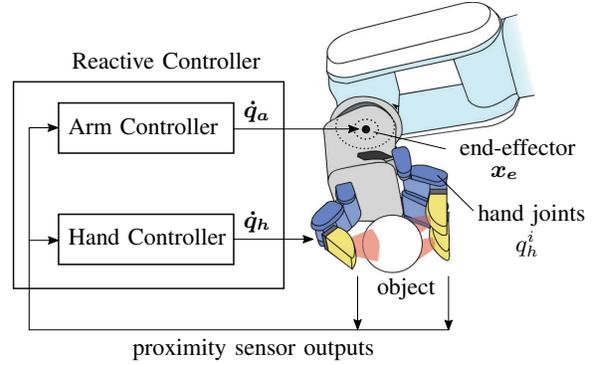
    \caption{Reactive control for preshaping is implemented by feeding back the outputs obtained from the proximity sensors mounted on a robot hand, i.\,e.\ its palm and fingers, to controllers that affect the pose $q_{h}^{i}$ of each finger and the end-effector $\bm{{x_{e}}}$.}
    \label{fig:PreshapingControl}
\end{figure}

In this section, we describe reactive preshaping using proximity sensor feedback, i.\,e.\ closed-loop control (\gls{ATII}). The concept of preshaping has been proposed in psychology in the context of studies on human grasping, e.\,g.\ in \cite{pellegrino1989timecourse}, before it was adopted in robotics. In robotics terms, preshaping describes the (preliminary) motions of adjusting finger joint poses $\bm{q_{h}}$ of a robot hand and end-effector pose $\bm{x_{e}}$ before grasping, as shown in Figs.~\ref{fig:Intro_Preshaping} and~\ref{fig:PreshapingControl}. Preshaping is usually based on visual cues, such as global object shape or the detection of affordances, as is the case for the human. However, proximity sensing opens an opportunity in robotics to implement this behavior by closing the perception loop to increase robustness and performance, i.\,e.\ an ad-hoc solution. In fact, the traditional, human-like preshaping can be considered to be a high-level behavior (\gls{BTII}) and is the result of \emph{grasp synthesis} or \emph{grasp planning}, which is an active research field~\cite{bohg2013data}. In contrast, preshaping using feedback from proximity sensors is often reflex-like, i.\,e.\ \gls{BTI}. We, therefore, call it \emph{reactive preshaping} (see also Fig.~\ref{fig:ApplicationsAndBehaviors}). Reactive preshaping has significant potential for \gls{HRI}, because the automatic adaption of the robot hand to the object pose generalizes naturally to the case when the object is not static, for instance during handover tasks.

Reactive preshaping enhances grasping robustness and performance, from reaching to grasping motions, for the following reasons (see also Fig.~\ref{fig:Intro_Preshaping}): 

\begin{itemize}
  \item The contact area after grasping is increased and the grasping becomes stable by aligning the normals of the object and fingertip surfaces before establishing contact. 
  \item ``[...] ensuring that the fingers contact the object simultaneously can improve the probability of successfully grasping the object''~\cite{Mayton2010}, i.\,e.\ unwanted object motion is avoided.
  \item A robot can move fast without moving or damaging an object because sensing is without contact.
  \item A robot can continuously adjust the end-effector and finger joint poses even when occlusion occurs in the vision sensor.
\end{itemize}

\subsubsection{Reactive Preshaping Control}
To implement reactive preshaping, poses or torques of finger joints are controlled directly based on the signals detected by proximity sensors on the surface of the finger, as illustrated by Fig.~\ref{fig:PreshapingControl}. An early contribution in this area is due to Espiau and Catros~\cite{espiau1980}, who show closed-loop control of a two-jaw gripper. Mayton et al.~\cite{Mayton2010} realized finger reactive preshaping using the Barrett Hand with mid/short-range electric field (capacitive) sensors. Each finger was controlled independently by PID control of the motor current in the hand using a target proximity sensor target value. They demonstrated reactive preshaping on a banana and a juice bottle on a table, as well as grasping these objects. The developed method also allowed the robust handover of objects with the human and detection of the co-manipulation state from the capacitive signals.

Hsiao et al.~\cite{Hsiao2009} proposed a reactive grasping controller, including a finger distance controller, with real-time calibration using optical proximity sensor outputs (see also Fig.~\ref{fig:Hsiao2009_final}). To detect the actual distance and posture of an object's surface, they proposed a calibration method based on a probabilistic model using fingertip positions and reflected light intensity values. The average distance sensing error was reported as $\SI{4}{\milli\meter}$, the posture error of pitch rotation was $\SI{5.3}{\degree}$, and the posture error of roll rotation was $\SI{17.7}{\degree}$ degrees for common objects. At the beginning of the procedure, the fingertips are controlled using the raw sensor output as a target value. The target value is then switched from raw value to estimated distance and posture once the estimated values are calculated. 

Escaida~Navarro et al.~\cite{Escaida2014b,Escaida2015b} installed capacitive tactile proximity sensors on a parallel jaw gripper, and they realized simultaneous control of six DoF of a two-jaw gripper based on a proximity-feedback control. In~\cite{Escaida2014b}, there are $2\!\times\!2$ sensor areas per finger of the gripper, and the posture and position information of the object are detected simultaneously by comparing the intensity of the sensor values in each area. Somewhat similar is the work of Guan et al.~\cite{Guan2012}, who use reactive preshaping in translation and orientation of a gripper to a pole the robot is climbing. In a sensor arrangement comparable to ~\cite{Escaida2014b,Escaida2015b}, Guo et al.\ also use a $2\!\times\!2$ of \gls{OBB} sensor arrangement for reactive preshaping in~\cite{Guo2015}.

\begin{figure}
    \centering
    \includegraphics[width=0.7\linewidth]{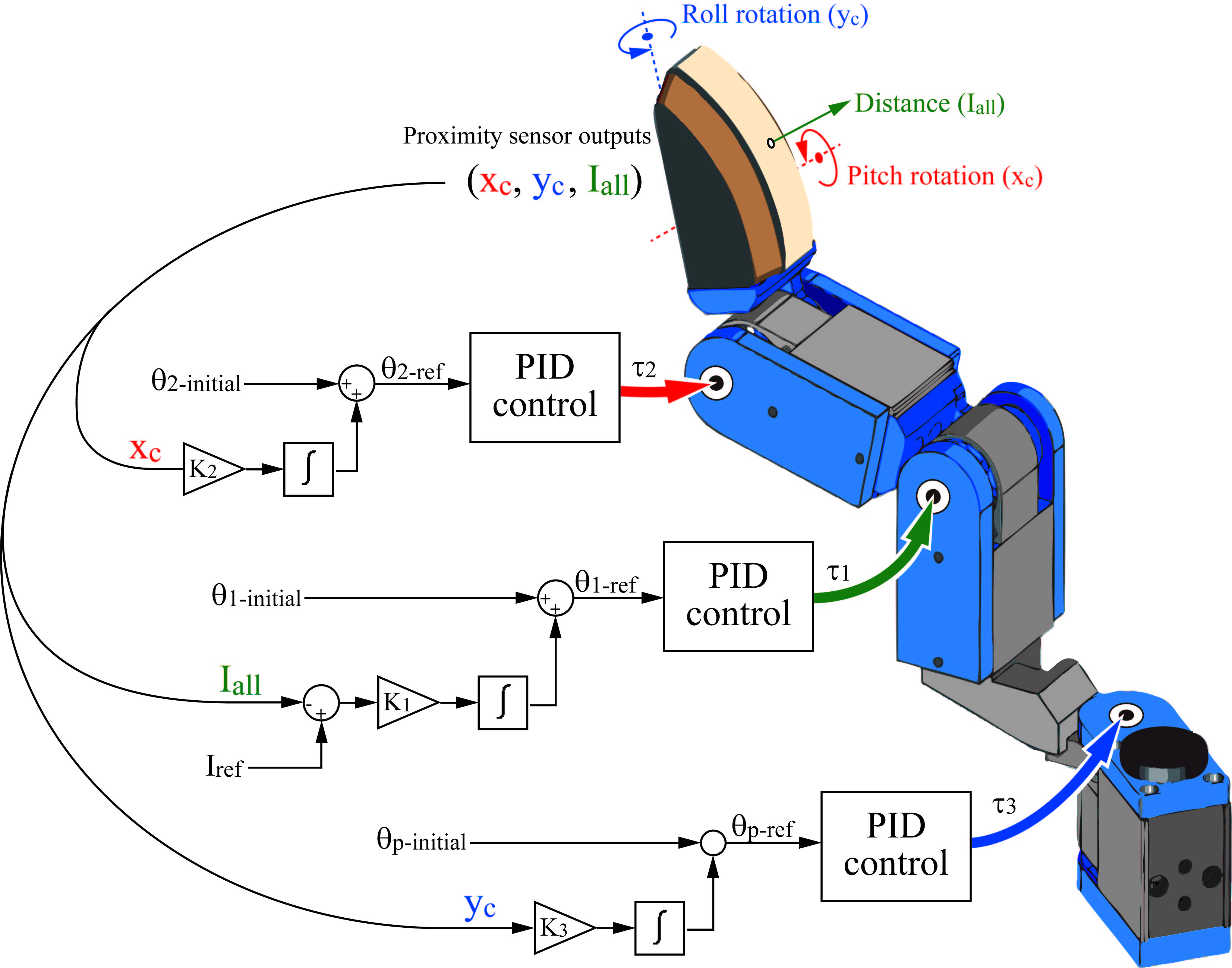}
    \caption{Independent joint angle control for preshaping using a high-speed proximity sensor as shown by Koyama et al.~\cite{koyama_ijrr2019}. (\copyright 2019 Sage Publications)}
    \label{fig:koyama2019_pc}
\end{figure}

Koyama et al.~\cite{koyama2013,koyama2016,koyama_ijrr2019} realized independent control of joint poses of an 8-DoF robot hand (three fingers) using a high-speed optical proximity sensor. They demonstrated reactive preshaping for an apple, a banana, and moving objects on a conveyor or during handover, with simple joint-angle controls (see Fig.~\ref{fig:koyama2019_pc}), updating the sensor value and control every $\SI{5}{\milli\second}$. In these preshaping controls~\cite{Mayton2010,koyama_ijrr2019}, the target value is set in advance from experimental data of an object set. 

Furthermore, Koyama et al.~\cite{koyama2015} realized velocity control of fingers without relying on a previously established calibration. The control uses \emph{\gls{TTC}} values calculated using proximity values from the sensor. \gls{TTC} is the remaining time until a collision between an object and the sensor. \gls{TTC} is a bio-inspired calculation that does not depend on the surface characteristics of an object. Therefore, the relative speed between the fingertip and an object can be controlled without prior reflectance data. 
The authors also realized high-speed catching of soft objects using a high-speed, high-precision proximity sensor in~\cite{koyama2018}. As the proposed design has impressive distance resolution, it is possible to utilize this value to estimate the contact condition of an object. Elastic pads of $\SI{3}{\milli\meter}$-thickness in front of the photodiode provide deformable spacers (see Fig.~\ref{fig:koyama2013-2019_optical}~b)). Any distance measurement that is equal or closer to the offset provided by the elastic pads is thus indicative of a contact situation. A contact situation with a soft object can be detected with very low contact force. The contact detection enabled the catching of very soft objects, namely a marshmallow and a paper balloon, with negligible deformation~\cite{koyama2019}. 

In~\cite{erickson2018tracking,erickson2019multidimensional} Erickson et al.\ apply the principles of reactive preshaping/contour following to the scenario of dressing and washing of patients. In~\cite{erickson2019multidimensional}, a $2\!\times\!3$ array on the end-effector is used to align the end-effectors distance and orientation to the patient's limbs, also following their contour. They show this is a viable approach to automated caregiving tasks (dressing, washing), as visual occlusions are amortized and the detection of the human is reliable.

\subsection{Higher Complexity Methods and Behaviors (Towards \gls{BTII})}
\label{subsec:highlevelApplications}

In this subsection, we discuss what contributions can be found in the literature regarding cognitive and model-based methods and behaviors (\gls{BTII}, see Sec.~\ref{subsec:ApplicationsAndBehaviorTypes} and Fig.~\ref{fig:ApplicationsAndBehaviors}). 

\subsubsection{Pre-touch Exploration (\gls{ATII})}

\begin{figure}
    \centering
    \includegraphics[width=0.7\linewidth]{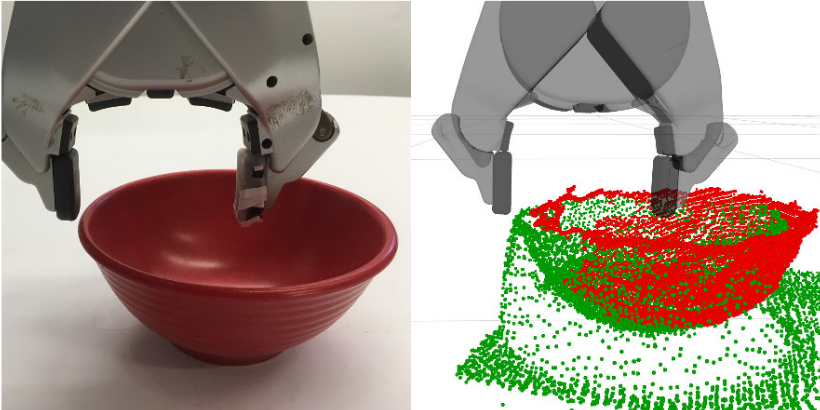}
    \caption{Complementing a pointcloud obtained by the MS-Kinect with \gls{OTOF} sensor inside the gripper PR-2, as shown by Lancaster et al.\ \cite{lancaster2017improved}. (\copyright 2017 IEEE)}
    \label{fig:Lancaster2017}
\end{figure}

Proximity perception opens the opportunity of aggregating information to an object model without mechanical contact. Pre-touch exploration means executing a systematic strategy using the robot's tool to acquire and aggregate pre-touch data into an object model. Jiang et al.\ present an exploration strategy for completing a point cloud obtained from an RGB-D camera in~\cite{Jiang2012,Jiang2013}. The object point clouds originating from these cameras are incomplete on two accounts, i.\,e.\ the occluded backside of the object and due to translucency. These perception gaps undermine grasp-planning algorithms that have to rely on object geometry knowledge. In their approach, the unknown regions of the objects are explored until stable grasp planning is possible. A similar approach is due to Maldonado et al.~\cite{Maldonado2012} who, in addition to completing point clouds for grasp planning tasks, take advantage of the imaging capability of a proposed sensor for the classification of surfaces (textures). In~\cite{Guo2015}, Guo et al.\ realized a reactive pre-touch control using an optical proximity sensor. First, in the control loop, the bounding box of the object was detected from Kinect point cloud information, and the initial grasping point was determined. Second, the object shape information was refined by detecting the edge of the object while tracing it with the proximity sensor. Finally, a better grasping position was determined by repeatedly executing grasp planning and pre-touch detection. The researchers realized grasping of tissue paper, a difficult task with only vision/depth and tactile sensor feedback.  In~\cite{lancaster2017improved}, Lancaster et al.\ study the use of deep learning to guide a proximity-based exploration strategy. It yields an improved object model as well as an improved estimate of the object's pose. Fig.~\ref{fig:Lancaster2017} shows an example of a point cloud that is being complemented by pre-touch exploration. In~\cite{patel_integrated_2018}, Patel et al.\ use their optical proximity sensors inside a two jaw gripper to scan objects by moving the gripper around them. The robot's kinematics allows the easy aggregation of the data to point cloud data. A similar result is shown by Markvicka et al.\ in~\cite{markvickawireless}, where they discuss the scanning of a model space shuttle with a robot hand with enough details to capture its most important features (fuselage, wings, etc.).

In~\cite{Escaida2014b}, Escaida Navarro et al.\ show object exploration as an application of proximity servoing. Edges can be explored continuously by adjusting the gripper pose as the exploration progresses. To obtain the precise location of corners, the strategy is complemented by the acquisition of tactile samples, iteratively, in the regions where corner candidates are detected in proximity mode. In~\cite{kaboli2017tactile} and~\cite{kaboli2018active}, Kaboli et al.\ use a multi-modal sensor skin (see~\cite{Mittendorfer2011}) to explore and classify objects according to haptic properties. The number and location of objects on a table in the workspace of the robot are determined by a Bayesian pre-touch exploration strategy. A similar approach for scanning a table workspace for graspable objects, using \gls{OTOF}, is followed by Yin et al.\ in~\cite{yin2020closing}. In \cite{MuehlbacherKarrer2015b}, M\"ulbacher-Karrer et al.\ investigate how capacitive sensing can be used to sense the fill state of bottles using a hand equipped with sensors. The fill state is explored by tilting the bottle with the hand, whereby the decision (full or empty) is decided in a Bayesian framework. Finally, in~\cite{palermo2020automatic} it is proposed to use proximity sensing is used for increasing robustness in crack detection, although the exploration procedure itself is contact-based.

\subsubsection{Bio-inspired Sensing and Behavior (\gls{ATI}, \gls{ATII})}
\label{subsubsec:BioInspired}

In robotics, the mimicry of the behavior of weakly electric fish has delivered interesting results, which can be considered to be an inspiration for proximity sensing in general. In~\cite{boyer2013underwater}, Boyer et al.\ study basic control laws of a cylindrical under-water probe, featuring capacitive-like sensing based on a dipole-type arrangement, i.\,e.\ a voltage imposed on the tail and current measured on the tip. A model for the electrosense yields justification for a set of basic, reflex-like control laws that govern the behavior of the probe (\gls{ATI}, \gls{BTI}). In a related work~\cite{bazeille2017model}, Bazeille et al.\ show results for recognition of elliptical objects (\gls{ATII}, \gls{BTII}). A sequence of measurements of an object is obtained as the probe passes by. The material properties (conductive or insulating) and geometrical properties (ellipse parameters, location, and orientation) are found in an optimization framework. The optimization uses a forward model of the electrosense to find the parameter set that best explains the measured sequence. Tackling a similar problem, Bai et al.\ use active alignment and machine learning to identify spheroids with a biomimetic probe~\cite{bai2015finding} (\gls{ATII}, \gls{BTII}). In 2013, Neveln et al.\ presented a survey paper on the subject of biomimetic robotics related to weakly electric fish that significantly goes beyond the scope of what we can summarize here~\cite{neveln2013biomimetic}. 

\subsubsection{Teleoperation and VR (\gls{ATI}, \gls{ATII})}
As discussed in Sec.~\ref{subsec:CollisionAvoidance}, teleoperation has been seen as an application of proximity sensing from early on \cite{Lumelsky1993,wegerif1993whole,Feddema1994}. In none of these approaches, however, did the authors rely on a master-device with force or tactile displaying capability. In more recent approaches, using haptic cues that are originating from proximity sensors has been investigated. As Huang et al. put it in \cite{huang2018visionless}: ``Thus it (the haptic feedback) provides the perceptual benefits of touch interaction to the operator, without relying on the negative consequences of the robot actually contacting unknown geometrical structures.'' In~\cite{huang2018visionless} by Huang et\ al., point cloud data collected from a finger-tip sensor is used to generate virtual fixtures. It is also shown that proximity sensing promotes the teleoperation-based exploration of moving objects. 

Stoelen et~al.~\cite{stoelen_adaptive_2016,stoelen_adaptive_2013} presented their approaches using whole-arm sensitive robots. The application is a teleoperated robot arm with shared autonomy, which is based on the signals of the proximity sensors. The authors use machine learning to enable the robot to predict collisions from proximity sensor values based on prior experience. Therefore, the velocity of the robot is limited and force feedback through a \emph{PHANTOM Omni} is provided to the operator in~\cite{stoelen_adaptive_2016}. After three days of experiments based on a virtual environment, the authors show that the completion time of tasks, as well as the workload estimated by the users, decreased using the aid of the controller. In further works, authors have extended the visual perception installed on end-effectors with capacitive proximity sensors. In~\cite{Escaida2015b}, Escaida Navarro et al.\  generate 6D force-feedback from the proximity signals to aid the user in exploration tasks and Alagi et al.\ use tactile feedback with spatial resolution to help users in detecting shape cues in~\cite{alagi2020}. Both works show how proximity sensing can close the perception gap caused by visual distortions or occlusions in teleoperation.

In~\cite{stetco2020b}, remote gesture-based control of a mobile manipulator, based on capacitive proximity sensors, was presented. Different operation modes such as control of the end-effector or the mobile platform were demonstrated. Advantages of the capacitive interface acting as a virtual 3D mouse to control the robot are the robustness of the sensor against water, occlusions or even objects covering parts of the sensor interface (this work does not feature force-feedback).

Finally, the new advances in augmented and virtual reality theologies provide a new kind of representation of proximity information. Beyond the visual augmentation, one can combine it with force, tactile, audio \cite{marquardt2018audio-tactile}, or even with transcutaneous electrical stimulation \cite{zhao2020electrically}, addressing different human sensation to increase the level of presence in teleoperation.

\subsubsection{Material Classification (\gls{ATII})}
\label{subsubsec:MaterialClassification}
Beyond object exploration based on its geometry, the internal properties, such as the material or electrical properties, are very valuable knowledge for reliable grasping robust object manipulation. For example, using capacitive sensing, the relative permittivity $\varepsilon_r$ of an object can be estimated. It is then possible to classify it according to its material. The exciter frequency dependency of $\varepsilon_r(\omega)$ can be then utilized to identify the material. Kirchner et al.\  presented an approach to identify material by performing multi-frequency capacitive sensing~\cite{kirchner2008capacitive}. The researchers drove the circuit with three different frequencies and were able to classify $7$ different materials. A similar approach was presented in \cite{ding2018capacitive}, driving the electrodes with $290$ exciter frequencies between \SI{10}{kHz} and \SI{300}{kHz} and analyzing both the amplitude and the phase of the corresponding signals. This method is also known as \emph{capacitive spectroscopy}, referring to the different exciter frequencies used to perform the measurements. Furthermore, in \cite{alagi2018material}, another approach for material recognition using the flexible spatial resolution of a capacitive sensor array was presented. The sensors were driven with two excitation frequencies at different electrode configurations, in which the size of the electrodes changed. Using \gls{GPR}, automated mapping of material layers for investigation of soil composition is possible~\cite{kouros20183d}. Based on pulse-echo ultrasound and optoacoustic effects, Fang et al. reported in \cite{fang2019toward} the feasibility of integrating optical and acoustical measurement systems into a fingertip of a robotic gripper. A preliminary study showed successful material classification with an accuracy over of $87\%$ for three materials (steel, rubber, and acrylic). It is worth mentioning that estimating material properties also plays a big role in the bio-inspired approaches previously addressed (Sec.~\ref{subsubsec:BioInspired}). 

\subsubsection{Tracking (\gls{ATI}, \gls{ATII})}

Profiting from distributed sensing, object tracking from proximity sensing streams also has been investigated to some extent in the literature. In~\cite{Petryk1996,Petryk1997}, Petryk and Buehler install four O-RLI-type sensors in a gripper and show that an extended Kalman filter can serve to track the 2D-position of a cylindrical object with respect to the gripper. Furthermore, the reflectance of the object is also estimated. The potential for \gls{ATII} is discussed. In~\cite{Escaida2013a}, the authors show an approach for tracking objects detected on a $3\!\times\!16$ array of capacitive sensors. The task is handled like an image processing problem. Also using a Kalman filter, the authors show the capability of tracking two hands and handling occlusions, i.\,e.\ targeting \gls{HRI} tasks (\gls{ATI}).

\subsection{Industrial Technologies and Solutions}
\label{subsec:IndustrialTechnologies}
In recent years, proximity sensing technology emerged on the market mainly driven by the industry to deploy collaborative robots in production lines, for instance in the automotive industry. The engineering and technology company BOSCH introduced \emph{BOSCH APAS} for flexible human-robot collaboration (\gls{HRC}), which is a mobile robot system for industrial applications. The sensor skin of the manipulator utilizes capacitive based proximity sensor technology (\gls{CM}) to enable safe \gls{HRC} in an industrial manufacturing environment~\cite{bosch_apas_nodate,Frangen2010}. Also recently, FOGALE Robotics~\cite{FOGALErobotics} presented a smart skin for robots based on capacitive proximity sensing technology (\gls{CSE}). The capacitive based multi-modal (tactile and proximity) robot skin reaches a sensor range of up to $\SI{300}{\milli\meter}$, where the electrodes are arranged in a matrix structure on the surface of the robot manipulator. In~\cite{mcolo2019obstacle}, the skin was utilized together with a control framework to avoid obstacles for \gls{HRI}. The KUKA system partner MRK-Systeme provides a sensor skin solely based on capacitive sensor technology for KUKA robots for industrial \gls{HRC} applications. In~\cite{hoffmann2016environment} MRK-Systeme presented capacitive based proximity perception for \gls{HRI} (\gls{CSE}) in industrial environments, utilizing a sensor front-end with an electrode configuration able to achieve up to $\SI{350}{\milli\meter}$ of sensing range on a KUKA KR6 manipulator.

\section{Future Perspectives: Grand Challenges}
\label{sec:FuturePerspectives}

In this section, we provide an outlook for the domain of human-centered proximity perception in robotics. We have highlighted what we think are \emph{grand challenges} at the end of each the three sub-sections. 

\subsection{Human-robot interaction in the industry and service domains}
\label{subsec:HRIInIndustryAndServiceRobotics}
Proximity perception technology is mature enough to be deployed under strict safety requirements, as discussed in Sec.~\ref{subsec:IndustrialTechnologies}. However, today, collaborative automation still struggles to be an interesting value proposition, i.\,e.\ providing a large increase in efficiency that will justify the cost of investing in this technology. A value proposition that is more likely to be attractive in the near future is that of \emph{fenceless automation}. Here, humans and robots coexist in the same space but do not necessarily share a task. Value is added for instance by the fact that real-estate on the shop-floor is saved or because robotic automation is now possible in spaces that were previously considered to be too small. Nonetheless, safety certification is still a major challenge that is preventing wide-spread use of proximity perception. Today, technologies such as the ones discussed in ~\ref{subsec:IndustrialTechnologies} need to be certified on a solution level. Certification at the modular sensor level is not yet possible. Therefore, technologies such as radar (see Sec.~\ref{subsec:Radar}) can be attractive alternatives for HRI in industrial automation. Radar is more likely to achieve a safety rating on modular level soon, also profiting from all the prior experience coming from the developments in autonomous driving.

Areas such as medical robotics face similar issues for commercialization. The process of certifying the solutions is long and costly. However, we think that the market for automated solutions in health care or service robotics based on proximity perception is there, especially in scenarios where the human is very close to the robot, preventing the sole use of cameras. In this survey, we discussed examples such as grasping of moving objects and handover~\cite{Mayton2010,koyama2016,koyama_ijrr2019}, dressing and washing of patient \cite{erickson2018tracking,erickson2019multidimensional}, or assistive robotics based on teleoperation \cite{stoelen_adaptive_2013,stoelen_adaptive_2016}. Overall, we think that medical and service robotics is a promising field for proximity perception, but more research is needed to establish use-cases and the corresponding methods before commercial exploitation appears on the horizon.Furthermore, the study of highly redundant robotic systems that use proximity sensing is scarce. Meanwhile, the exploitation of kinematic redundancy based on proximity sensor feeds is natural, as discussed in Secs.~\ref{subsubsec:EarlyJacobian} and \ref{subsubsec:RecentJacobian}. This includes the use of proximity perception in unusual areas, such as on the legs or feet of robots. More research, like the one done in the group of Prof.\ Cheng~\cite{cheng2019comprehensive}, is needed. In summary, we can say the following about the challenges in these domains:
\begin{itemize}
    \item Lowering the difficulties in achieving safety ratings for proximity sensing technologies will significantly expand the market for fenceless or collaborative automation. An effective procedure for safety certification, elaborated by industrial stakeholders and certification organizations, is needed.
    \item Modularized technologies, such as radar chips, can have significant advantages in a certification procedure because solutions can  be based on safety rated components.
    \item In the service, medical and active assisted living domains proximity perception allows for new ways of interaction between humans and robots yet more research is needed before commercially viable applications are established.
    \item A better understanding of how highly redundant robotic systems can profit from proximity perception is needed.
    \item The potential of proximity perception for increasing robustness of grasping of moving objects, for instance during handover tasks or in teleoperation, needs to be established further, for example with conclusive user studies.
\end{itemize}

\subsection{Cognitive Robotics}
\label{subsec:CognitiveRobotics}

\begin{figure}
    \centering
    \includegraphics[width=0.95\linewidth]{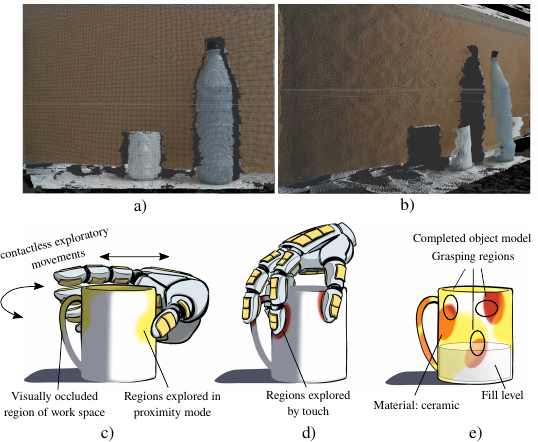}
    \caption{Proximity perception can play an important role in multi-modal exploration: a) Shows a typical scene with objects as captured by an RGBD-camera. b) A perspective shift reveals considerable gaps in the visual perception, having occluded elements that could be explored contactless and contact-based exploration steps, c) and d) respectively. A multi-modal cognitive model controls the exploration and aggregates the information to the object model, which includes geometric, material property, fill-level information as well as potential grasping regions obtained from the exploration steps in c) and d). This aggregated information is illustrated in e). The yellow and red colors on the mug (in c)-d)) illustrate regions that are extracted from the proximity and touch-based perception, respectively. }
\label{fig:futureperspectives}
\end{figure}

One of the hallmarks of cognitive robotics is active perception. It is the paradigm in which perception is enabled by purposeful actions, i.\,e.\ exploration). The active perception principle has been studied in many robotics research papers, as discussed in a survey by Bohg et al.~\cite{bohg2017interactive}. Machine learning is essential in active perception for tasks such as object pose estimation or scene labeling, which can be based on proximity perception, e.\,g.~\cite{lancaster2017improved}. Currently, a trend in active perception is the combined perception of vision and touch. We think that extending this trend to include proximity sensing will be a relevant research question in the near future. Therefore, we will see haptic exploration strategies that will be able to explore occluded regions of the workspace combining touch-based and touch-less exploratory movements. We have illustrated this possible multi-modal exploration approach in Fig.~\ref{fig:futureperspectives}. Some aspects of it have already been addressed in the literature discussed in this paper, e.\,g.~\cite{Maldonado2012,MuehlbacherKarrer2015b,varley2017shape,lancaster2017improved}. Another important domain in this area is the multi-modal modeling of the human for safe interaction and collaboration. In summary, regarding the major challenges we can say that:
\begin{itemize}
    \item As we mentioned throughout the paper, there is a need for proximity sensing technologies that can easily be combined with vision and/or touch so they can be deployed together. This will make it attractive to include proximity sensing in established and novel active perception/cognitive approaches.
    \item It can be expected that the trend regarding sim-to-real learning is going to prevail for the next years. Therefore, it is a challenge to implement realistic simulation models for the different measurement principles discussed in Sec.~\ref{sec:PhysicalWorkingPrinciples}. In most cases, current approaches to model them are not viable in terms of their temporal performance. Overcoming this is an important challenge.
\end{itemize}

\subsection{Soft Robotics}
\label{subsec:SoftRobotics}

We think that a good portion of the topics relevant for proximity perception will eventually transfer to Soft Robotics. In this way, soft manipulators equipped with proximity sensors can perform collision avoidance in an analogous way as described in Sec.~\ref{subsec:CollisionAvoidance}, i.\,e.\ respecting task hierarchies. Similarly, soft robots will be able to execute reactive preshaping, grasping, and exploration tasks using proximity sensors. Arguably more so than in other areas, in Soft Robotics it is of interest to study how to purposefully engage in contacts to achieve the desired task. Proximity perception can help in finding and reaching desired contact states. Overall, the support of simulation frameworks for model-based control and sensing, such as SOFA~\cite{navarro2020model}, will also be relevant. Therefore, regarding the major challenges in this domain, we can say that:
\begin{itemize}
    \item New control methods must be found for soft robots to integrate the information provided by the proximity sensors and adapt the actuation strategy adequately.
    \item Integration of proximity sensors in deformable structures represents an important challenge. A significant cross-talk between the global deformation and the detection of tactile and proximity events has to be expected. As stated in the previous section, appropriate models for proximity sensors are needed for sim-to-real learning or for interpreting their signals correctly under deformation using interactive simulations~\cite{shin_effect_2018,navarro2020model}. 
    \item Except for capacitive sensing, the realization of proximity sensors having deformable or stretchable sensing elements is challenging. Shrinking the size of rigid components, such as ICs, within a stretchable substrate may still allow the realization of deformable circuits, as described e.\,g. in \cite{nagels2018silicone,hellebrekers2018liquid}.
\end{itemize}

\section{Summary and Conclusions}
\label{sec:SummaryAndConclusions}
In this paper, we have given an overview of the main aspects of proximity sensing in today's robotic landscape. Considering that the field has not had a significant formalization over the years, we provide a basic scheme for categorization of the robotic applications and technologies based on proximity sensors and we propose a set of traits that characterize proximity sensors in human-centered robotics. We give an account of the existing technologies and the main measurement principles reported by authors since the early 1970s and have organized the technologies in Table~\ref{tab:ComparisonYear}, including characteristics such as sensing range, update rate, sensing element size, etc. We then proceeded to detail how the technologies have been used for implementing applications such as collision avoidance and human-robot interaction (\gls{ATI}) as well as preshaping and grasping (\gls{ATII}). We start with the seminal developments of the early years in these domains and cover the progress up to today (2021). The tight integration into the sensory-motor functionality has received constant attention over the years in order to realize highly reactive behavior of robotic systems (\gls{BTI}). Meanwhile, as the area of robotics progresses as a whole, we report that more and more approaches begin to have cognitive aspects in them (\gls{BTII}). \gls{BTII} includes areas such as teleoperation, where the human is in the loop, autonomous object exploration and even bio-inspired approaches that mimic weakly electric fish. 

Finally, Sec.~\ref{sec:FuturePerspectives} is dedicated to summarizing our projections for the field regarding the grand challenges we have identified. We think that as the technology of proximity sensors is reaching the maturity to coexist with tactile and visual perception in terms of integration, costs, and norm conformity, it will be adopted for a variety of solutions, especially those involving interaction with humans.

\section*{Acknowledgment}

This work was supported by the Region Hauts-de-France, the  project COMOROS (ANR-17-ERC2-0029), the European Regional Development Fund (ERDF) and the project Inventor (I-SITE ULNE, le programme  d’Investissements d’Avenir, M\'etropole Europ\'eenne de Lille). 

This work has received funding from the "K\"arntner Wirtschaftsf\"orderung Fonds" (KWF) and the "European Regional Development Fund" (EFRE) within the CapSize project 26616/30969/44253. 

This work also was supported by the Federal Ministry of Education and Research (Bundesministerium f\"ur Bildung und Forschung, BMBF) within the project (Verbundprojekt-Nr.: 16SV7823K:). 

 The authors would like to thank cartoonist Adriana Filippini for the illustrations. 

\ifCLASSOPTIONcaptionsoff
  \newpage
\fi



%
\bibliographystyle{IEEEtran}
\bibliography{ReviewProximityPerception}

%

\begin{IEEEbiography}[{\includegraphics[width=1in,height=1.25in,clip,keepaspectratio]{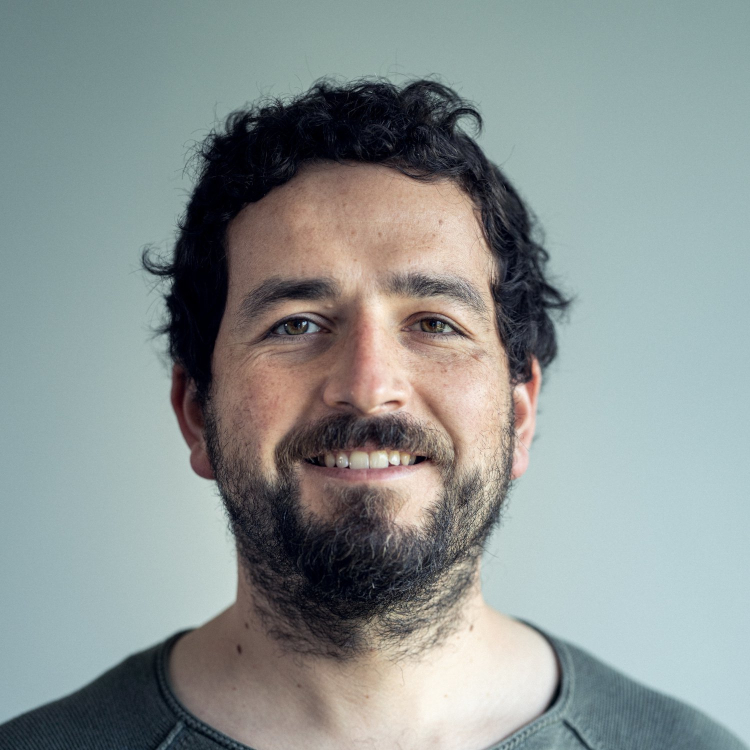}}]{Stefan Escaida Navarro}
received his diploma in computer science (Dipl.-Informatiker) from Karlsruhe Institute of Technology (KIT), Germany in 2010. He obtained is PhD degree (Dr.-Ing.) from the same institution in 2016. During his doctoral phase he researched on haptic object recognition and grasping as well as the technology and the applications for capacitive tactile proximity sensors, such as preshaping, collision avoidance and teleoperation. He is currently a postdoctoral researcher at Inria Lille - Nord Europe, France, where he is working on model-based force and shape sensing for Soft Robotics. 
\end{IEEEbiography}

\begin{IEEEbiography}[{\includegraphics[width=1in,height=1.25in,clip,keepaspectratio]{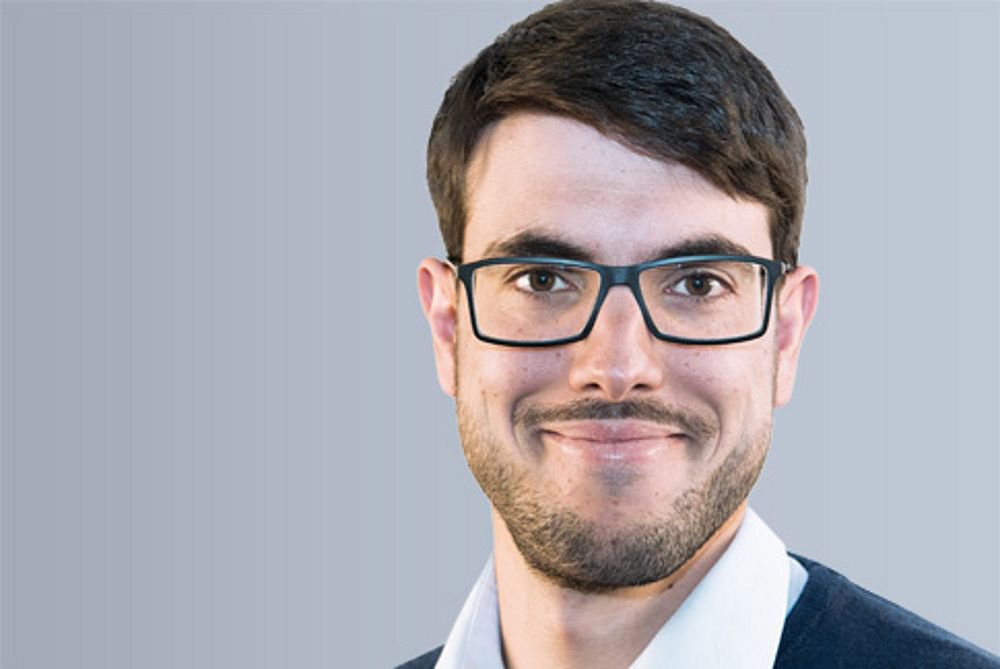}}]{Stephan M\"{u}hlbacher-Karrer}
received the Dipl.Ing. degree in telematics from the Graz University of Technology (TUG), Graz, Austria, in 2012, where he focused on Autonomous Robots and Telecommunications. In 2017 he received his Dr. tech. degree in information and communications engineering from the Alpen-Adria-Universit\"{ä}t Klagenfurt, Klagenfurt, Austria, focusing on capacitive sensing for robotic applications.
From 2012 to 2013, he was with Gigatronik Ingolstadt GmbH, Gaimersheim, Germany, working
for Audi Electronics Venture GmbH, Gaimersheim, where he was involved in the pre-development for autonomous driving. Since 2017 he is a senior researcher with JOANNEUM RESEARCH ROBOTICS, Institute for Robotics and Mechatronics in Klagenfurt Austria. His current research interests include near field sensor technology, signal processing, and autonomous robots.
\end{IEEEbiography}

\begin{IEEEbiography}[{\includegraphics[width=1in,height=1.25in,clip,keepaspectratio]{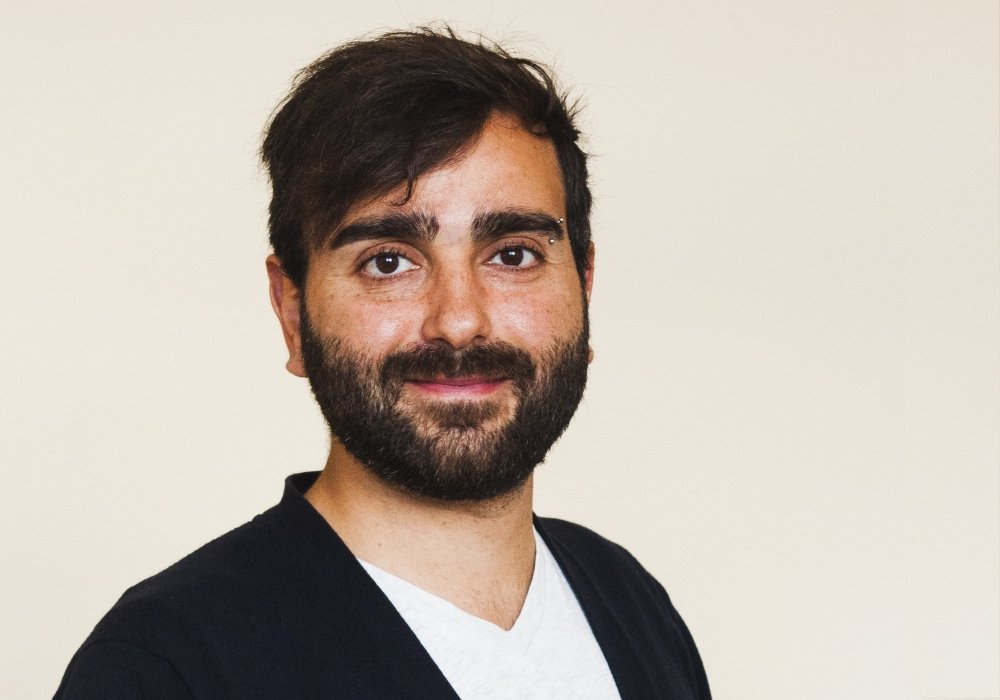}}]{Hosam Alagi}
received his M.Sc. degree in electrical engineering and information technology at the Karlsruhe Institute of Technology in Germany with a specialization in System-on-Chip. Since August 2015 he is working as a research assistant at the Institute for Anthropomatics and Robotics (IAR) at Intelligent Process Automation and Robotics Lab (IPR). Focusing on human-robot interaction, he develops capacitive multi-modal sensors and evaluates methods that allow robots to perceive their near environment. Within the scope of the PhD Studies both the sensors and sensor applications are being implemented, addressing the industrial and domestic domains.
\end{IEEEbiography}

\begin{IEEEbiography}[{\includegraphics[width=1in,height=1.25in,clip,keepaspectratio]{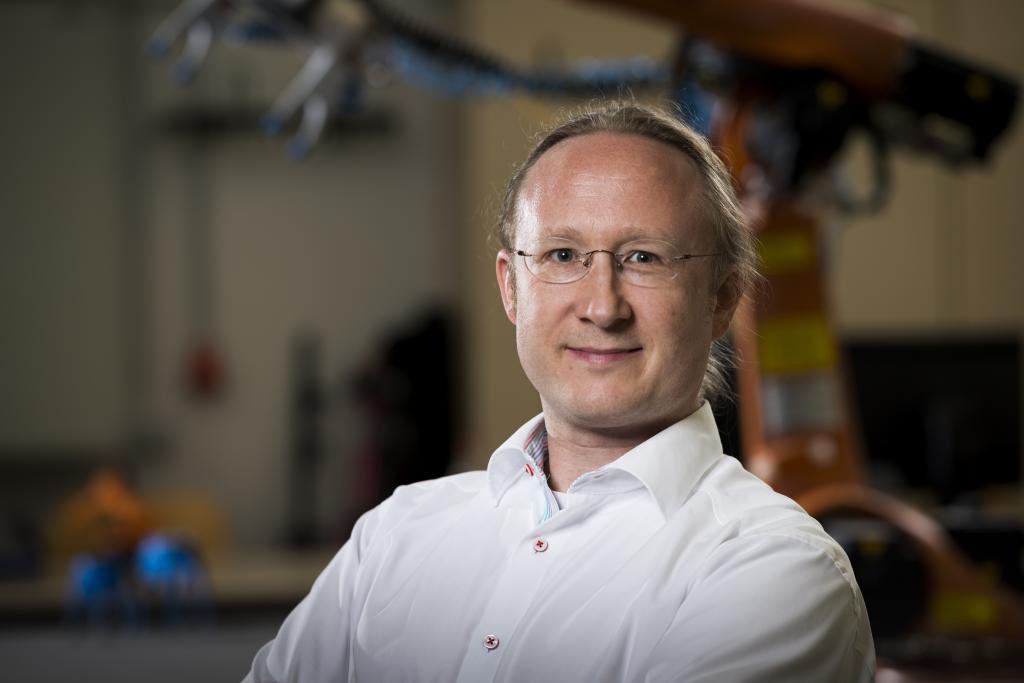}}]{Bj\"{o}rn Hein}
studied electrical engineering with a focus on control theory and received his PhD in 2003 concerning automatic collision-free path planning at the Karlsruhe Institute of Technology (KIT). May 2010 he finished his post-doctoral lecture qualification for computer science with stress on human-robot interaction. 2012-2018 he was professor for Interaction Technologies for Robot Systems at the Institute for Anthropomatics and Robotics (IAR) - Intelligent Process Automation and Robotics Lab (IPR). Currently he holds a professorship at the University of Applied Sciences in Karlsruhe. His research focus comprises: algorithms for collision free motion planning and path optimization, methods for intuitive and automatic programming of robots, human-robot interaction, novel sensor-technologies for enhancing capabilities of robot systems.
\end{IEEEbiography}

\begin{IEEEbiography}[{\includegraphics[width=1in,height=1.25in,clip,keepaspectratio]{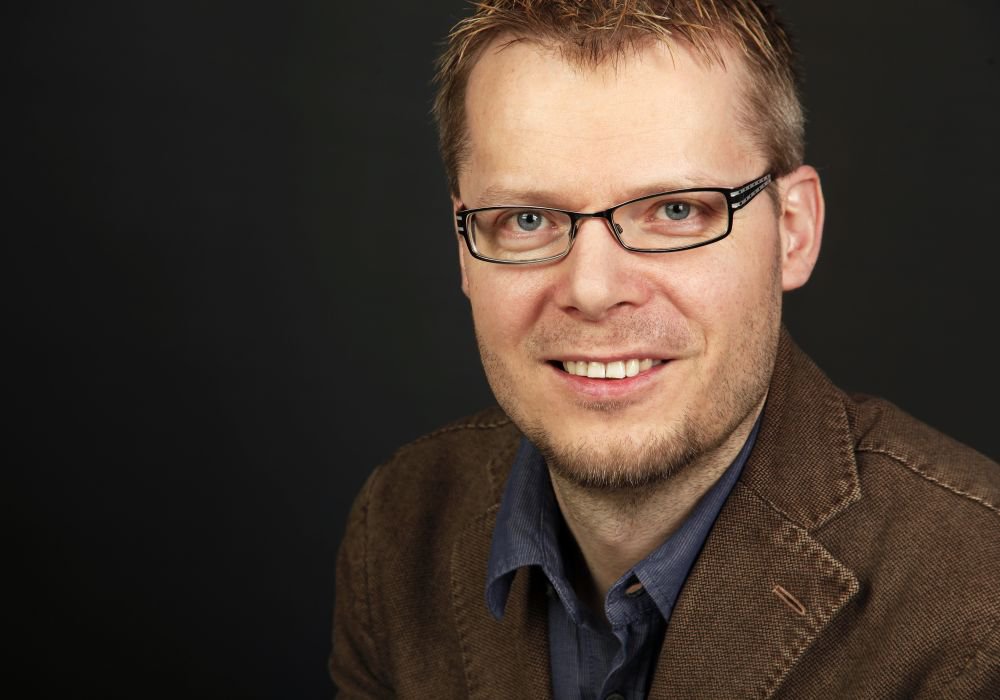}}]{Hubert Zangl} received the Dipl.Ing. degree in telematics, the Dr. Tech. degree in electrical engineering, and the Venia Docendi degree in sensors and instrumentation from the Graz University of Technology (TUG), Graz, Austria, in 2001, 2005, and 2009, respectively.
From 2010 to 2013, he was an Associate Professor of Sensors and Instrumentation with the Institute of Electrical Measurement and Measurement Signal Processing, TUG. Since 2013, he has been a Professor chairing Sensors and Actuators with the Institute of Smart System Technologies, Alpen-Adria-Universität Klagenfurt, Klagenfurt, Austria. 
His current research interests include the design and optimization of smart sensors and actuators, robustness, and reliability of sensors and actuators, sensor signal processing, autarkic wireless sensors, and energy harvesting with target applications in the field of IoT and robotics.
\end{IEEEbiography}

\begin{IEEEbiography}[{\includegraphics[width=1in,height=1.25in,clip,keepaspectratio]{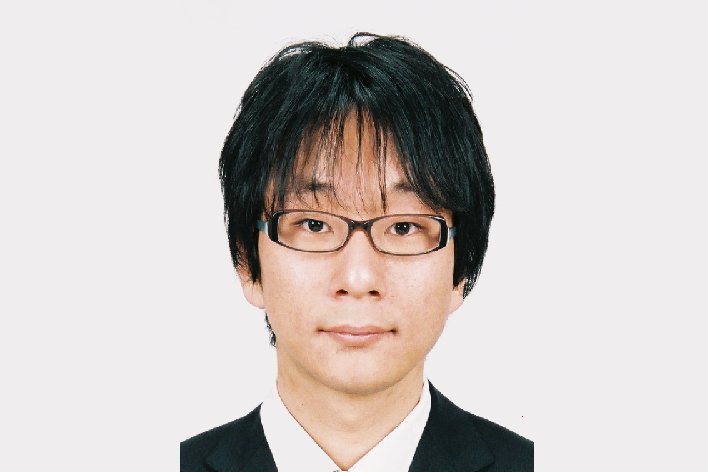}}]{Keisuke Koyama} received the B.E., M.E. and Ph.D. in Eng. degrees in mechanical engineering and intelligent Systems from the University of Electro-Communications (UEC), 
  in 2013, 2015 and 2017, respectively.   He was a research fellow of Japan Society for the Promotion of Science (JSPS) from 2015 to 2017. 
  He researched on high-speed proximity sensor for pre-grasping and integrated control of multi-degree-of-freedom robot hand and arm. 
  From 2017 to 2019, he was project Assistant Professor, Department of Information Physics and Computing, Graduate School of Information Science and Technology, The University of Tokyo. 
  Since 2019, he has been Assistant Professor, Department of Systems Innovation, Graduate School of Engineering Science, Osaka University. 
  And, he has been Visiting Researcher, Department of Information Physics and Computing, Graduate School of Information Science and Technology, The University of Tokyo. 
  His current research focuses on high-speed, high-precision proximity sensing for high-speed robotic manipulation and assembly. 
\end{IEEEbiography}

\begin{IEEEbiography}[{\includegraphics[width=1in,height=1.25in,clip,keepaspectratio]{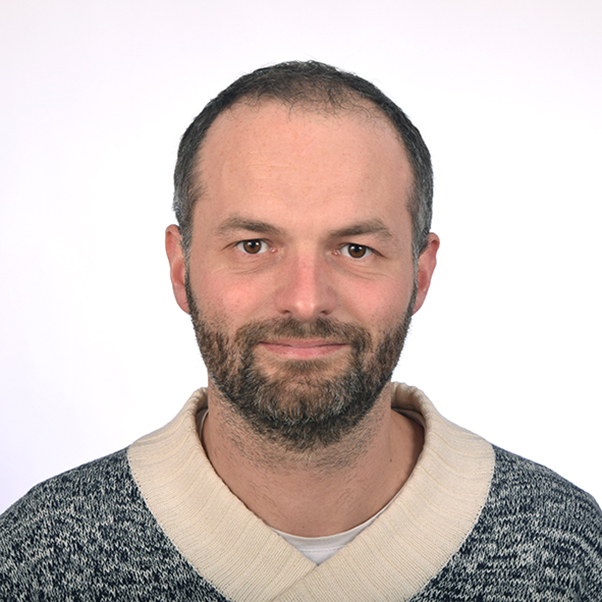}}]{Christian Duriez} is Research Director at Inria Lille - Nord Europe.
He received the engineering degree from the Institut Catholique d’Arts et Métiers of Lille, France and a PhD degree in robotics from University of Evry and CEA in France in 2004. He had a postdoctoral position at the CIMIT SimGroup in Boston. He arrived at INRIA in 2006 and worked on interactive simulation of deformable objects and haptic rendering with focus on surgical simulation. He is now the head of DEFROST team, created in January 2015. In 2018 he was invited researcher at Stanford University. His research topics are Soft Robot models and control, Fast Finite Element Methods, simulation of contact response, new algorithms for haptics… He participated to the creation of the open-source SOFA framework. He was also one of the founders of the start-up company InSimo.
\end{IEEEbiography}

\begin{IEEEbiography}[{\includegraphics[width=1in,height=1.25in,clip,keepaspectratio]{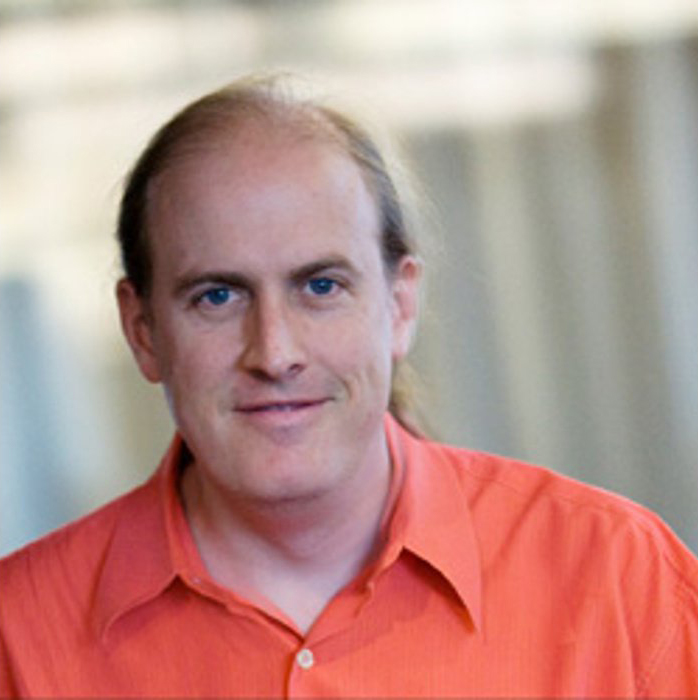}}]{Joshua R. Smith} is the Milton and Delia Zeutschel Professor, jointly appointed in the Allen School of Computer Science and Engineering, and the Department of Electrical Engineering, at the University of Washington, Seattle. He received BA degrees in Computer Science and Philosophy from Williams College in 1991, an MA in Physics from the University of Cambridge in 1997, and SM and PhD degrees in 1995 and 1999 from the MIT Media Lab's Physics and Media Group, where he began working on Electric Field Sensing for input device applications. He began applying these techniques to robotics while he was a researcher at Intel Labs Seattle, between 2004 and 2010. In recent years his group at UW has developed a number of new approaches to proximity perception for robotics.

\end{IEEEbiography}






\newpage
\markboth{APPENDIX to Proximity Perception in Human-centered Robotics - A Survey on Sensing Systems and Applications}%
{Shell \MakeLowercase{\textit{et al.}}: Bare Demo of IEEEtran.cls for IEEE Journals}
\onecolumn
\setcounter{page}{1}

\begin{center}
\Huge{Appendix} \\ \vspace{.5cm}
\end{center}


    
\printglossaries

\clearpage
\newpage

\begin{landscape}
\begin{center}
\large{Overview - Sorted by year - Table I}
\end{center}
\tiny
\begin{doclongtable}{p{0.3cm}p{1.3cm}p{0.8cm}p{0.8cm}p{1.1cm}p{1.3cm}p{1.1cm}p{1.3cm}p{2.1cm}p{1.3cm}p{2.7cm}p{3.5cm}p{1.3cm}} 
\toprule
{} & Measurement Principle &                            Reference &  Year & Reported Min. Working Distance [mm] &                           Reported Max. Range [mm] &              Field of View [degree] &                            Measurement Rate [Hz] & Sensing Element Dimension [$mm$ / $mm^2$ / $mm^3$] & Multiple-Obstacles & Commercially Available Core Components &                                     Categorization &                 Basis reference \\
\midrule
\endhead
\midrule
\multicolumn{13}{r}{{Continued on next page}} \\
\midrule
\endfoot

\bottomrule
\endlastfoot
1    &                 O-Tri &          \cite{johnson_optical_1973} &  1973 &                                   2 &                                                 20 &                                 n/s &                                              n/s &                                              3x5x5 &     not considered &                                     no &                   AT-II, Manipulation and Grasping &                               - \\
2    &                 O-Tri &           \cite{Lewis1973PlanningCF} &  1973 &                                   2 &                                                 20 &                                 n/s &                                              n/s &                                              3x5x5 &     not considered &                                     no &                     AT-II, Grasping, Teleoperation &     \cite{johnson_optical_1973} \\
4    &                 O-Tri &                   \cite{fuhrman1984} &  1984 &                                 n/s &  v1: focal point 4.5  +-0.56 cm, v2: focal poin... &                                  13 &                                             1000 &                                                110 &                 no &                                     no &  AT-II, measure distance, orientation and curva... &                               - \\
7    &                 O-RLI &              \cite{cheung1988motion} &  1988 &                                 n/s &                                                250 &                                  60 &                                    16 whole skin &                               min. one sensor pair &                yes &                                     no &                          AT-I, Collision Avoidence &               \cite{Cheung1992} \\
6    &                   C-M &                   \cite{Yamada1988}  &  1988 &                                  40 &                                                100 &                                 n/s &                                              n/s &                                             80x210 &         not tested &                                     no &                            AT-I Obstacle Detection &                               - \\
8    &                 O-RLI &         \cite{cheung1989development} &  1989 &                                 n/s &                                                250 &                                  60 &                                    16 whole skin &                               min. one sensor pair &                yes &                                     no &                          AT-I, Collision Avoidence &               \cite{Cheung1992} \\
9    &                  A-US &                     \cite{Dario1990} &  1990 &                                 n/s &                                            200-300 &                                 180 &                                              n/s &                                          ~20x20x20 &                n/s &                                    n/s &                                              AT-II &                               - \\
10   &                   C-M &    \cite{vranish_capaciflector_1991} &  1991 &                                   0 &              304.8 human and alu, 127 grafite lead &                                 n/s &                                              n/s &                                        355,6x190,5 &                 no &                                    n/s &                          AT-I, Collision Avoidence &                               - \\
13   &                 O-RLI &                    \cite{Cheung1992} &  1992 &                                 n/s &                                                250 &                                  60 &                                    16 whole skin &                               min. one sensor pair &                yes &                      OD8810 and SFH205 &                          AT-I, Collision Avoidence &                               - \\
14   &                 O-RLI &             \cite{wegerif1992sensor} &  1992 &                                 n/s &                                                200 &                                 n/s &                                    30 whole skin &                                                n/s &                n/s &                                     no &                          AT-I, Collision Avoidence &                               - \\
12   &                   C-M &                     \cite{Novak1992} &  1992 &                                 n/s &                                                400 &                                 n/s &                                              100 &                                                n/s &         not tested &                                    n/s &                          AT-I, Collision Avoidance &                               - \\
16   &                 O-RLI &              \cite{wegerif1993whole} &  1993 &                                 n/s &                                                200 &                                 n/s &                                    30 whole skin &                                                n/s &                n/s &                                     no &                          AT-I, Collision Avoidance &        \cite{wegerif1992sensor} \\
17   &                 O-RLI &                  \cite{Lumelsky1993} &  1993 &                                 n/s &                                                250 &                                  60 &                                    16 whole skin &                               min. one sensor pair &                yes &                                     no &                          AT-I, Collision Avoidence &               \cite{Cheung1992} \\
15   &                   C-M &            \cite{novak1993collision} &  1993 &                                 n/s &                                                400 &                                 n/s &                                              100 &                                                n/s &         not tested &                                    n/s &                          AT-I, Collision Avoidance &                \cite{Novak1992} \\
19   &                   C-M &                  \cite{Feddema1994}  &  1994 &                                 n/s &                                                400 &                                 n/s &                                              100 &                                                n/s &         not tested &                                    n/s &                          AT-I, Collision Avoidance &                \cite{Novak1992} \\
18   &                  A-US &                     \cite{Nunes1994} &  1994 &                                 n/s &                                                n/s &                                 n/s &                                              n/s &                                                n/s &                n/s &                                    n/s &                           AT-II, Contour Following &                               - \\
20   &                 O-RLI &                    \cite{Petryk1996} &  1996 &                                   5 &                                                 75 &                          100 (+-50) &                                              300 &                                            ~5x5x13 &                 no &                                    STM &                   AT-II, Manipulation and Grasping &                               - \\
22   &                 O-RLI &                     \cite{bonen1997} &  1997 &                                 n/s &                                                 40 &                           60 (+-30) &                                            10000 &                                                ~32 &                 no &                                     no &                   AT-II, Manipulation and Grasping &                               - \\
24   &                 O-RLI &                    \cite{Petryk1997} &  1997 &                                 n/s &                                                200 &                          130 (+-65) &                                              n/s &                                2020-08-07 00:00:00 &                 no &                                     no &                AT-II, Object Tracking for Grasping &                               - \\
23   &            A-US,O-RLI &               \cite{tamasy1997smart} &  1997 &               25 (A-US), 50 (O-RLI) &                           3000 (A-US), 450 (O-RLI) &                                 n/s &                           40 (A-US), 100 (O-RLI) &                                       63.5x44.5x20 &                yes &                                     no &                                  AT-I, Sensor Skin &                               - \\
25   &                  O-BB &                 \cite{Teichmann2000} &  2000 &                                 n/s &                                                n/s &                                 n/s &                                              n/s &                                                n/s &                 no &                                     no &                AT-II, BT-I, Sensing for Preshaping &                               - \\
27   &                  C-SE &                    \cite{Stiehl2006} &  2006 &                                   0 &                                                n/s &                                 n/s &                                              200 &                                               15x5 &                 no &                                MC33794 &                                  AT-I, Sensor Skin &                               - \\
29   &                 O-RLI &                    \cite{Walker2007} &  2007 &                                   0 &                                                 50 &                                 360 &                                               48 &                                                  1 &                yes &                                     no &                      AT-I, BT-I, Contour Following &                               - \\
28   &                   C-M &                     \cite{smith2007} &  2007 &                                 n/s &                                                 50 &                                 180 &                                              n/s &                                               ~50  &                 no &                                     no &                      AT-II, Sensing for Preshaping &                               - \\
31   &                   C-M &        \cite{kirchner2008capacitive} &  2008 &                                 n/s &                                                n/s &                                 n/s &                                              500 &                                                n/s &                n/s &                                     no &                        AT-II, Material recognition &                               - \\
32   &                   C-M &                   \cite{Wistort2008} &  2008 &                                 n/s &                                                190 &                                 n/s &                                               30 &                                                ~50 &                n/s &                                     no &                          AT-II, Proximity Servoing &                               - \\
33   &                 O-Tri &                       \cite{Tar2009} &  2009 &                                  10 &                                                 60 &                                 n/s &                                              224 &                                           90 x 90  &                yes &                               TCRT1000 &                          AT-I, AT-II, Sensor Array &                               - \\
36   &                 O-RLI &                     \cite{Hsiao2009} &  2009 &                                   2 &                                                 40 &    4 alnaog sensors per finger = 90 &                                              n/s &                                          6x3.7x3.7 &                n/s &                        Vishay TCND5000 &                     AT-II, Preshaping and Grasping &                               - \\
35   &                   C-M &                       \cite{Lee2009} &  2009 &                                 n/s &                                                170 &                                 n/s &                                              n/s &                                              22x22 &                 no &                                     no &                                AT-I, AT-II, Sensor &                               - \\
38   &             C-SE, C-M &                   \cite{Goeger2010a} &  2010 &                                 n/s &                                                100 &                                 n/s &                                           30-100 &                                              40x40 &                yes &                                     no &  Preshaping, Grasping, Exploration, Teleoperati... &                               - \\
37   &                   C-M &                    \cite{Mayton2010} &  2010 &                                   0 &                                                150 &                                 n/s &                    20 (update cycle for control) &                           Fingertip of Barret Hand &         not tested &                                     no &    AT-II, Preshaping, Grasping and Co-manipulation &                               - \\
40   &                   C-M &                   \cite{Frangen2010} &  2010 &                                 n/s &                                                n/s &                                n/s  &                                              n/s &                                                n/s &                yes &              BOSCH (complete solution) &                                       AT-I, Sensor &                               - \\
42   &                   C-M &                    \cite{George2010} &  2010 &                                   0 &                                                160 &                                 180 &                                            25000 &                                            220x150 &                 no &                                     no &                          Automotive Seat Occupancy &                               - \\
43   &                 O-RLI &              \cite{Mittendorfer2011} &  2011 &                                 n/s &                                                  3 &                           n/s (180) &                                             1000 &                                                3x4 &                n/s &                                 GP2S60 &              AT-I, AT-II, Modular Multi-Modal Skin &                               - \\
44   &                   C-M &                \cite{boyer2011model} &  2011 &                                   0 &                                                ~25 &                                 360 &                                            ~1000 &                                         whole body &                n/s &                                     no &                           AT-I, AT-II, bioinspired &                               - \\
46   &                   C-M &                   \cite{Schlegl2011} &  2011 &                                   0 &                                               2000 &                                 360 &                                              n/s &                                          30 x 570  &                yes &                                 AD7143 &                                 Automotive parking &                               - \\
48   &                 O-RLI &                 \cite{Maldonado2012} &  2012 &                                   1 &                                                 10 &                              narrow &                                              n/s &                                              15x15 &                 no &                              ADNS-9500 &  AT-II, Slip Detection, Object Rreconstruction,... &                               - \\
49   &                  A-US &                      \cite{Guan2012} &  2012 &                                  20 &                                                400 &                                  15 &                                               40 &                                      45 x 20 x 15  &                 no &                                HC-SR04 &                        AT-II legged climbing robot &                               - \\
47   &                   A-S &                     \cite{Jiang2012} &  2012 &                                 n/s &                                2020-04-03 00:00:00 &                                 n/s &                                               20 &                                                  6 &                 no &                                    yes &       AT-II, Reactive Grasping, Object Exploration &                               - \\
57   &                 O-Tri &            \cite{ceriani2013optimal} &  2013 &                                 200 &                                               1500 &                                 n/s &                                 250 (20 sensors) &                                           ~20 x 40 &         not tested &                       Sharp GP2Y0A02YK &                          AT-I, Collision Avoidance &                               - \\
52   &                 O-RLI &         \cite{stoelen_adaptive_2013} &  2013 &                                  10 &                                                400 &                                  10 &                                              ~50 &                                          simulated &                yes &                       TCND5000,GP2D120 &  AT-I, AT-II, Teleoperation, Shared Autonomy, S... &                               - \\
59   &                 O-RLI &                    \cite{koyama2013} &  2013 &                                   0 &                                                 50 &                                  90 &                                             1000 &                                           22x24x40 &                n/s &                              EE-SY1200 &                                  AT-II, Preshaping &                               - \\
54   &             C-SE, C-M &                    \cite{Goeger2013} &  2013 &                                 n/s &                                                100 &                                 n/s &                                           30-100 &                                            40 x 40 &                yes &                                     no &                   AT-II, Multi-Modal Sensor System &                               - \\
55   &                  C-SE &                  \cite{Escaida2013a} &  2013 &                                 n/s &                                                100 &                                 n/s &                                           30-100 &                                            40 x 40 &                yes &                                     no &  AT-II, Hand and Object Tracking, Collision Pre... &               \cite{Goeger2013} \\
51   &                   C-M &                  \cite{Schlegl2013b} &  2013 &                                   0 &                                             $>$100 &                                 180 &                                             6250 &                                      1500x150x0.35 &                 no &                                     no &           AT-I, BT-I, Reactive Collision Avoidance &                               - \\
60   &                   C-M &           \cite{boyer2013underwater} &  2013 &                                   0 &                                                ~25 &                                 360 &                                            ~1000 &                                         whole body &                n/s &                                     no &                         AT-I, AT-II, Bio-inspired  &           \cite{boyer2011model} \\
61   &                   A-S &                     \cite{Jiang2013} &  2013 &                                 n/s &                                2020-04-03 00:00:00 &                                 n/s &                                               20 &                                                  6 &                 no &                                    yes &      AT-II,  Reactive Grasping, Object Exploration &                \cite{Jiang2012} \\
63   &                 O-Tri &            \cite{avanzini2014safety} &  2014 &                                 200 &                                               1500 &                                 n/s &                                 250 (20 sensors) &                                             ~20x40 &         not tested &                       Sharp GP2Y0A02YK &                          AT-I, Collision Avoidance &       \cite{ceriani2013optimal} \\
62   &                  C-SE &                  \cite{Escaida2014b} &  2014 &                                 n/s &                                                100 &                                 n/s &                                           30-100 &                                              40x40 &                yes &                                     no &        AT-II, Preshaping, Grasping and Exploration &               \cite{Goeger2013} \\
65   &                   C-M &                    \cite{Zhang2014}  &  2014 &                                   0 &                                                 90 &                                 n/s &                                              n/s &                                                5x5 &                yes &                                     no &                                      Sensor System &                               - \\
66   &                 O-RLI &        \cite{konstantinova2015force} &  2015 &                                   0 &                                                 20 &                                 n/s &                                            ~1000 &                                             ~3 x 1 &                 no &        KEYENCE fiber optical converter &                   AT-II, Multi-Modal Sensor System &                               - \\
67   &                 O-RLI &                    \cite{koyama2015} &  2015 &                                   0 &                                                 50 &                                  90 &                                             1000 &                                           22x24x40 &                n/s &                              EE-SY1200 &                                              AT-II &               \cite{koyama2013} \\
72   &                 O-RLI &                  \cite{Hasegawa2015} &  2015 &                                 n/s &                                                300 &                                 n/s &                                             1000 &                                            100x100 &     not considered &                                     no &                                        AT-I, AT-II &                               - \\
73   &                  O-BB &                       \cite{Guo2015} &  2015 &                                   0 &                                              $>$84 &                                 n/s &                                              n/s &                                          32x19x6.5 &                 no &                                      - &                     AT-II, Preshaping and Grasping &                               - \\
68   &                   C-M &                \cite{bai2015finding} &  2015 &                                   0 &                                               ~250 &                                 360 &                                              n/s &                                         whole body &                 no &                                     no &                          AT-I, AT-II, Bio-Inspired &                               - \\
69   &                   C-M &                  \cite{Escaida2015b} &  2015 &                                 n/s &                                                100 &                                 n/s &                                           30-100 &                                              40x40 &                yes &                                     no &                  AT-II, Teleoperation, Exploration &               \cite{Goeger2013} \\
70   &                   C-M &         \cite{MuehlbacherKarrer2015} &  2015 &                                   5 &                                                 50 &                       Cross Section &                                              n/s &                                            200x200 &                yes &                                     no &           AT-I, BT-II Object detection, tomography &                               - \\
71   &                   C-M &        \cite{MuehlbacherKarrer2015b} &  2015 &                                   0 &                                                n/s &                                 180 &                             quasi-simultaneously &                                     Meka H2 Finger &                 no &                                     no &  At-II, BT-II Active Object Categorization, Gra... &                               - \\
83   &                     R &                     \cite{watts2016} &  2016 &                                 n/s &                         500 (for given resolution) &                       58 (E) 57 (H) &                                              n/s &                           13,2 (antenna diameter)  &                yes &                                     no &                          BT-II, Object exploration &                               - \\
76   &                 O-RLI &    \cite{konstantinova2016fingertip} &  2016 &                                   0 &                                                 20 &                                 n/s &                                            ~1000 &                                             ~3 x 1 &                 no &        KEYENCE fiber optical converter &                   AT-II, Multi-Modal Sensor System &   \cite{konstantinova2015force} \\
77   &                 O-RLI &                    \cite{koyama2016} &  2016 &                                   0 &                                                 50 &                                  90 &                                             1000 &                                           22x24x40 &                n/s &                              EE-SY1200 &                     AT-II, Preshaping and Grasping &               \cite{koyama2013} \\
81   &                 O-RLI &         \cite{stoelen_adaptive_2016} &  2016 &                                  10 &                                                400 &                                  10 &                                              ~50 &                                          simulated &                yes &                       TCND5000,GP2D120 &  AT-I, AT-II, Teleoperation, Shared Autonomy, S... &    \cite{stoelen_adaptive_2013} \\
78   &               C-SE, I &                 \cite{han2016highly} &  2016 &                                 n/s &                                                150 &                                 n/s &                                                5 &                                             30x30  &                n/s &                                     no &             AT-I, AT-II, Multi-Modal Sensor System &                               - \\
80   &             C-SE, C-M &                    \cite{Alagi2016a} &  2016 &                                   0 &                                                100 &                                 n/s &                                           22-380 &                                     20x20 to 40x40 &  yes, but only x/y &                                     no &             AT-I, AT-II, Multi-Modal Sensor System &                               - \\
75   &                  C-SE &                  \cite{Escaida2016a} &  2016 &                                 n/s &                                                100 &                                 n/s &                                           30-100 &                                              40x40 &                yes &                                     no &                            AT-I, Contour Following &               \cite{Goeger2013} \\
79   &                  C-SE &       \cite{hoffmann2016environment} &  2016 &                                   0 &                                                350 &                                 n/s &                                               40 &                                                n/s &                yes &        MRK-Systeme (complete solution) &                                         AT-II, HRI &                               - \\
74   &                   C-M &        \cite{MuehlbacherKarrer2016a} &  2016 &                                   0 &                                             50/100 &                                 180 &                                              n/s &                                            100x150 &                yes &                                     no &  AT-II, Gesture Control, Grasping \& Object Man... &                               - \\
82   &                   C-M &                  \cite{Escaida2016b} &  2016 &                                   0 &                                                100 &                                 n/s &                                           22-380 &                                     20x20 to 40x40 &  yes, but only x/y &                                     no &                     AT-II, Preshaping and Grasping &               \cite{Alagi2016a} \\
86   &           O-ToF, C-SE &           \cite{yang_pre-touch_2017} &  2017 &                                  10 &                                                100 &  180 for whole fipgertip (6 module) &                                               30 &                                     4.8 x 2.8 x 1  &                n/s &                                VL6180x &                          AT-II, Object Exploration &                               - \\
85   &                 O-ToF &         \cite{lancaster2017improved} &  2017 &                                   0 &                                                 70 &                                  42 &                                               10 &                                     4.8 x 2.8 x 1  &         not tested &                     VL6180x ToF sensor &                     AT-II, Preshaping and Grasping &      \cite{yang_pre-touch_2017} \\
84   &                 O-RLI &             \cite{kaboli2017tactile} &  2017 &                                 n/s &                                                  3 &                           n/s (180) &                                             1000 &                                                3x4 &                n/s &                                 GP2S60 &                                       AT-II, BT-II &         \cite{Mittendorfer2011} \\
88   &               C-SE, I &              \cite{nguyen2017highly} &  2017 &                                 n/s &                                                150 &                                 n/s &                                                5 &                                             30x30  &                n/s &                                     no &             AT-I, AT-II, Multi-Modal Sensor System &            \cite{han2016highly} \\
87   &                  A-US &           \cite{steckel2017acoustic} &  2017 &                                 n/s &                                               6000 &                                 180 &                                             \~30 &                                          5.7 x 4.6 &                yes &                                     no &                                  AT-II, Navigation &                               - \\
96   &                     R &                  \cite{Flintoff2018} &  2018 &                                 n/s &                                                n/s &                                 n/s &                                              n/s &                                                n/s &                yes &                                    yes &                                    AT-II, Grasping &                               - \\
97   &                 O-Tri &                    \cite{koyama2018} &  2018 &                                2.85 &                                                 20 &                                  90 &                                             1000 &                                   18 x 28.5 x 38.5 &         not tested &                  VSMY1850, TEMD7500X01 &                             AT-II Dynamic grasping &                               - \\
98   &                 O-ToF &        \cite{hellebrekers2018liquid} &  2018 &                                   5 &                                                200 &                                   - &                                               30 &                                     4.8 x 2.8 x 1  &                n/s &                                VL6180x &                          AT-II, Object Exploration &                               - \\
100  &                 O-ToF &           \cite{huang2018visionless} &  2018 &                                   0 &                                                 70 &                                  42 &                                               10 &                                     4.8 x 2.8 x 1  &         not tested &                               VL6180x  &                               AT-II, Teleoperation &      \cite{yang_pre-touch_2017} \\
91   &          O-RLI, O-ToF &             \cite{sasaki2018robotic} &  2018 &                                   0 &                                                150 &                                 n/s &                                        1000, 100 &                     3.2 x 1.9 x 1.1, 4.8 x 2.8 x 1 &                 no &                    EE-SY 1200, VL6180x &                     AT-II, Preshaping and Grasping &                               - \\
90   &                 O-RLI &              \cite{kaboli2018active} &  2018 &                                 n/s &                                                  3 &                           n/s (180) &                                             1000 &                                                3x4 &                n/s &                                 GP2S60 &                          AT-II, Object Exploration &         \cite{Mittendorfer2011} \\
93   &                 O-RLI &           \cite{hughes_robotic_2018} &  2018 &                                   5 &      200 possible, 60-70 in their work due to PDMS &                                  60 &                                  20 (whole skin) &                                                4x4 &         not tested &                        Vishay VCNL4010 &  AT-I, AT-II, Preshaping, Grasping and Gesture ... &                               - \\
94   &                 O-RLI &         \cite{patel_integrated_2018} &  2018 &                                   5 &      200 possible, 60-70 in their work due to PDMS &                                  60 &                                  20 (whole skin) &                                                4x4 &         not tested &                        Vishay VCNL4010 &  AT-I, AT-II, Preshaping, Grasping and Gesture ... &                               - \\
95   &                  C-SE &          \cite{erickson2018tracking} &  2018 &                                   0 &                                                100 &                                 n/s &                                              200 &                                       115 x 85 x 1 &                n/s &                                 MPR121 &              AT-II, Contour Following, Health Care &                               - \\
89   &                   C-M &            \cite{ding2018capacitive} &  2018 &                                 n/s &                                                500 &                                 180 &  Capacitive: 1000, ToF: limited bei the i2C 400k &                                            40x40x4 &                yes &                                     no &                        AT-II, Material Recognition &         \cite{verellen2020high} \\
92   &                   C-M &             \cite{alagi2018material} &  2018 &                                   0 &                                                100 &                                 n/s &                                           22-380 &                                     20x20 to 40x40 &  yes, but only x/y &                                     no &                        AT-II, Material Recognition &               \cite{Alagi2016a} \\
99   &                  A-US &    \cite{marquardt2018audio-tactile} &  2018 &                                 n/s &                                                n/s &                                 n/s &                                              n/s &                                                n/s &                n/s &                                    n/s &                               AT-II, Teleoperation &                               - \\
104  &                     R &                    \cite{Geiger2019} &  2019 &                                 300 &                                                n/s &                                 n/s &                                              n/s &                             300 (waveguide length) &                yes &                                     no &                                               AT-I &                               - \\
110  &                 O-Tri &                    \cite{koyama2019} &  2019 &                                2.85 &                                                 20 &                                  90 &                                             1000 &                                   18 x 28.5 x 38.5 &         not tested &                  VSMY1850, TEMD7500X01 &                     AT-II, Preshaping and Grasping &               \cite{koyama2018} \\
102  &                 O-ToF &                 \cite{Lancaster2019} &  2019 &                                   0 &                                                 70 &                                  42 &                                               10 &                                     4.8 x 2.8 x 1  &         not tested &                     VL6180x ToF sensor &                   AT-II, Multi-Modal Sensor System &      \cite{yang_pre-touch_2017} \\
108  &                 O-RLI &        \cite{cheng2019comprehensive} &  2019 &                                 n/s &                                                  3 &                           n/s (180) &                                             1000 &                                                3x4 &                n/s &                                 GP2S60 &                    AT-I, AT-II, Whole-Body Control &         \cite{Mittendorfer2011} \\
109  &                 O-RLI &               \cite{koyama_ijrr2019} &  2019 &                                   0 &                                                 50 &                                  90 &                                             1000 &                                           22x24x40 &                n/s &                              EE-SY1200 &                     AT-II, Preshaping and Grasping &               \cite{koyama2013} \\
101  &           C-SE, O-ToF &                     \cite{tsuji2019} &  2019 &                                  10 &                                                100 &                       180 for whole &                                               30 &                                     4.8 x 2.8 x 1  &                n/s &                                VL6180x &                             AT-I, Multi-Modal Skin &                               - \\
103  &                  C-SE &  \cite{erickson2019multidimensional} &  2019 &                                   0 &                                                n/s &                                 n/s &                                              100 &                                            30 x 30 &                 no &                             Teensy 3.2 &              AT-II, Contour Following, Health Care &                               - \\
106  &                  C-SE &             \cite{mcolo2019obstacle} &  2019 &                                   0 &                                                300 &                                 n/s &                                              125 &                              50 x 50 to 100 x 100  &                yes &    FOGALE Robotics (complete solution) &                          AT-I, Collision Avoidance &                               - \\
111  &                  C-SE &                    \cite{Faller2019} &  2019 &                                   0 &                                                 50 &                                 180 &                                              n/s &                                             5 x 5  &                 no &                                     no &  AT-II, Grasping in Harsh environments, Industr... &                               - \\
107  &            C-M, O-ToF &                  \cite{ding2019with} &  2019 &                                 n/s &                                                500 &                                 180 &  Capacitive: 1000, ToF: limited bei the i2C 400k &                                            40x40x4 &                yes &                                     no &                          AT-I, Collision Avoidance &       \cite{ding2018capacitive} \\
105  &                  A-US &                \cite{fang2019toward} &  2019 &                                   0 &                                                 ~8 &                                 n/s &                                              n/s &                                          ~30x14x14 &                 no &                                     no &                        AT-II, Material Recognition &                               - \\
113  &                     R &                    \cite{stetco2020} &  2020 &                                 n/s &                                            $>$1200 &                                 160 &                                               40 &                                               9x13 &                yes &                                    yes &                                        AT-I, BT-II &                               - \\
118  &          O-Tof, O-RLI &             \cite{markvickawireless} &  2020 &                                   5 &                                                200 &                                   - &                                               30 &                                     4.8 x 2.8 x 1  &                n/s &                      VL6180x, MAX30105 &                          AT-II, Object Exploration &                               - \\
116  &                 O-ToF &                \cite{yin2020closing} &  2020 &                                   5 &                                                200 &                                   - &                                               30 &                                     4.8 x 2.8 x 1  &                n/s &                                VL6180x &                          AT-II, Object Exploration &   \cite{hellebrekers2018liquid} \\
119  &                 O-RLI &          \cite{zhao2020electrically} &  2020 &                                 n/s &                                                n/s &                                 n/s &                                              n/s &                                   7.1 x 2.75 x 2.7 &                 no &        Broadcom Limited, HSDL-9100-021 &  AT-II, Teleoperation, VR, Transcutaneous Elect... &                               - \\
115  &           C-SE, O-ToF &            \cite{tsuji2020proximity} &  2020 &                                   0 &                                                350 &                                 180 &                                               50 &                                              27x27 &                 no &                                     no &                                         AT-II, HRI &                               - \\
117  &               C-SE, I &                \cite{nguyen2020skin} &  2020 &                                   0 &                                                300 &                                n/s  &                                              100 &                                   100 x 100 x 2.75 &                n/s &                                     no &             AT-I, AT-II, Multi-Modal Sensor System &            \cite{han2016highly} \\
114  &                  C-SE &                     \cite{alagi2020} &  2020 &                                   0 &                                                100 &                                 n/s &                                           22-380 &                                     20x20 to 40x40 &  yes, but only x/y &                                     no &             AT-II, Teleoperation, Tactile Feedback &               \cite{Alagi2016a} \\
121  &                  C-SE &             \cite{poeppel2020robust} &  2020 &                                   0 &                                                350 &                                 180 &                                            \~108 &                                         \~175 x 80 &                 no &                                 AD7147 &                          AT-I, Collision Avoidance &  \cite{hoffmann2016environment} \\
112  &              C-M, ToF &             \cite{ding2020collision} &  2020 &                                 n/s &                                                500 &                                 180 &  Capacitive: 1000, ToF: limited bei the i2C 400k &                                            40x40x4 &                yes &                                     no &                           AT-I Collision Avoidance &       \cite{ding2018capacitive} \\
120  &                  A-US &              \cite{verellen2020high} &  2020 &                                 n/s &                                               6000 &                                 180 &                                             \~30 &                                          5.7 x 4.6 &                yes &                                     no &                                  AT-II, Navigation &                               - \\
\caption{Overview - Sorted by year \label{tab:ComparisonYear}}
\end{doclongtable}

\end{landscape}
\tableofcontents


\end{document}